\documentclass[runningheads]{llncs}

\usepackage{eccv}

\usepackage{eccvabbrv}

\usepackage{wrapfig}
\usepackage{graphicx}
\usepackage{booktabs}
\usepackage{tabularray}
\UseTblrLibrary{booktabs}
\usepackage{hyperref}
\usepackage{url}
\usepackage{booktabs}
\usepackage{multirow}
\usepackage[table]{xcolor}
\usepackage{float}
\usepackage{enumitem}
\usepackage{booktabs}
\usepackage{cellspace}
\usepackage{tabularx}
\usepackage[accsupp]{axessibility}  
\usepackage[numbers]{natbib} 

\usepackage{hyperref}
\usepackage{orcidlink}

\begin{document}

\title{Taming Text-to-Sounding Video Generation via Advanced Modality Condition and Interaction}
\titlerunning{Taming Text-to-Sounding Video Generation}

\author{%
Kaisi Guan\textsuperscript{*1,2},
Xihua Wang\textsuperscript{*1},
Zhengfeng Lai\textsuperscript{$\dagger$2},
Xin Cheng\textsuperscript{1},
Peng~Zhang\textsuperscript{2},
Xiaojiang Liu\textsuperscript{2},
Ruihua Song\textsuperscript{$\ddagger$1},
Meng Cao\textsuperscript{2}\\[0.5em]
\textsuperscript{1}Renmin University of China \quad
\textsuperscript{2}Apple \\
\email{guankaisi@ruc.edu.cn}
}

\authorrunning{K. Guan et al.}

\institute{}   

\maketitle

\footnotetext[1]{$^*$Equal contribution. Work done during an internship at Apple.}
\footnotetext[2]{$^\dagger$Project lead.}
\footnotetext[3]{$^\ddagger$Corresponding author.}
\begin{figure}[h]
    \centering
\includegraphics[width=\linewidth]{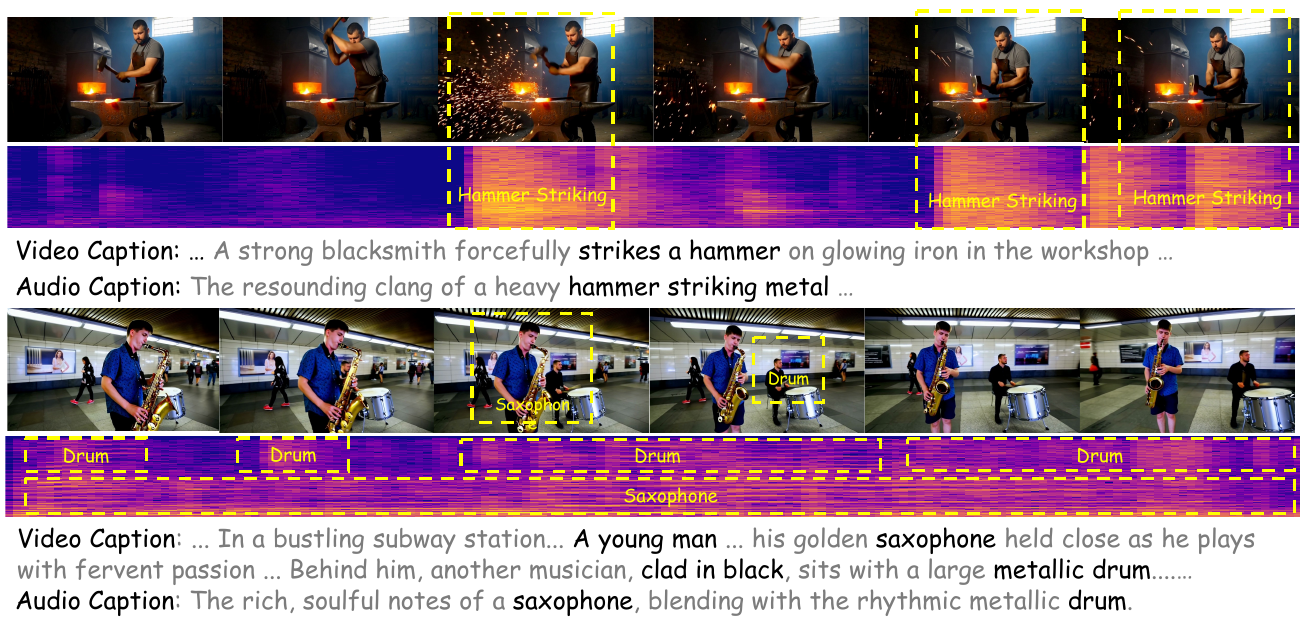}
    \caption{Examples of sounding videos generated by our BridgeDiT model, showcasing high quality, temporal synchronization, and text alignment. 
    Our method generates high-fidelity video frames and detailed audio spectrograms that remain faithful to the given text prompts. 
    Critically, as highlighted in the dashed boxes, the generated audio and video are precisely synchronized, demonstrating strong temporal coherence between visual events and their corresponding sounds. 
}
\label{fig:teaser}
\end{figure}
\vspace{-1.3cm}
\begin{abstract}
This study focuses on Text-to-Sounding-Video (T2SV) generation, which aims to generate a video with synchronized audio from text, with both modalities aligned to the text conditions. Despite progress in joint audio-video training, two critical challenges remain: (1) text conditioning is a bottleneck---shared captions ($T_V$ = $T_A$) trigger modal interference, while a gap persists between dense training captions and concise inference user prompts, and (2) the optimal fusion mechanism for cross-modal feature interaction remains unclear. To address the first challenge, we first propose the Cross-Referential Rewriter (CRR) caption framework, a dual-agent pipeline where a Semantic Checker extracts grounded Semantic Anchors and a Cross-Modal Rewriter generates disentangled caption pairs ($T_V$ and $T_A$), eliminating modal interference and bridging the training-inference gap via prompt expansion. For the second, we introduce BridgeDiT, a dual-tower diffusion transformer that employs Dual Cross-Attention (DCA) as a bidirectional bridge between video and audio streams, which we show through systematic comparison to be the optimal fusion strategy for the dual-tower paradigm. Extensive experiments on three benchmarks, supported by human evaluations, demonstrate state-of-the-art results on most metrics. Comprehensive ablation studies further validate each component and offer key insights for future T2SV systems. The codes and models are available at \href{https://bridgedit-t2sv.github.io/}{\nolinkurl{https://bridgedit-t2sv.github.io/}}.
\end{abstract}


\section{Introduction}
Human perception is inherently multi-sensory, with vision and sound tightly coupled.  Generating videos with synchronized audio from text (Text-to-Sounding-Video, T2SV) represents a crucial step toward world-modeling. Recent years have witnessed rapid progress in Text-to-Video (T2V)~\citep{blattmann2023stablevideodiffusionscaling,sora2024,opensora,kong2024hunyuanvideo,wan2025wan} and Text-to-Audio (T2A)~\citep{liu2023audioldm,huang2023makeanaudio,wang2025klingaudio,evans2025stableaudioopen} generation. With these unimodal capabilities becoming increasingly mature, the community naturally shifts attention to the more challenging task of T2SV~\citep{tang2023codi,liu2024syncflow,ishii2024ssvg,liu2025javisdit,weng2025mtv}.
Prior strategies for T2SV suffer from critical limitations. Generating video and audio independently fails to achieve temporal synchronization. Pipelined methods (e.g., T$\rightarrow$V$\rightarrow$A or T$\rightarrow$A$\rightarrow$V) attempt to address this, but suffer from error accumulation---the second-stage model (Video-to-Audio~\citep{wang2024tiva,cheng2025lova,cheng2025mmaudio,wang2025klingaudio} or Audio-to-Video~\citep{jeong2023tpos,yariv2024avalign,linz2024avsync}), trained only on ground-truth data, cannot correct first-stage errors and often amplifies them. To overcome this, research has shifted toward joint generation, where both modalities are synthesized simultaneously. The single-tower paradigm~\citep{ruan2022mmdiffusion,tang2023codi,sun2024mmldm,wang2024avdit,zhao2025uniform} learns the joint distribution from scratch but is data-intensive and difficult to optimize. The dual-tower architecture~\citep{ishii2024ssvg,liu2024syncflow,liu2025javisdit,wang2025jointdit,weng2025mtv} has thus emerged as the dominant approach, leveraging pretrained T2V and T2A backbones connected by a lightweight interaction module. Despite its promise, two fundamental challenges remain:

\textit{C1. The Conditioning Problem}: Previous dual-tower methods~\citep{liu2024syncflow,liu2025javisdit,zhao2025uniform,weng2025mtv} employ a shared caption ($T_V = T_A$) for backbones pretrained on distinct modalities, causing semantic interference. Moreover, a significant training-inference gap exists: models are trained on dense captions, but users provide concise prompts at inference, creating a distribution shift that degrades performance.

\textit{C2. The Interaction Problem}: The interaction module exchanges information between the two towers, but its optimal design remains unsolved. The core challenge is enabling effective bidirectional feature exchange to achieve both semantic and temporal synchronization.

We address the conditioning problem with the Cross-Referential Rewriter (CRR) caption framework. CRR is a dual-agent pipeline: a Semantic Checker cross-references raw captions from video and audio LLMs to extract grounded Semantic Anchors---filtering hallucinations and semantic conflicts---and a Cross-Modal Rewriter generates dense, disentangled captions ($T_V$, $T_A$) strictly from these anchors. At inference, the same pipeline expands concise user prompts into modality-specific descriptions matching the training distribution, bridging the training-inference gap without complex prompt engineering.
For the interaction problem, we propose BridgeDiT, a dual-tower architecture with Dual Cross-Attention (DCA) as its interaction module. Through systematic comparison with Full-Attention, Additive Fusion, and unidirectional alternatives under identical conditions, we identify DCA as the optimal fusion strategy: it preserves each tower's feature space while enabling symmetric, bidirectional information exchange, achieving strong synchronization without compromising generation quality.
Extensive experiments on three benchmarks demonstrate state-of-the-art performance, supported by human evaluations. Detailed ablation studies validate the necessity of each component---Semantic Anchors, prompt expansion, and bidirectional fusion---and provide practical insights for T2SV system design. In summary, our contributions are:
\begin{itemize}
    \item The CRR caption framework, a dual-agent pipeline that resolves the conditioning problem through grounded Semantic Anchors, modality-pure caption generation, and training-inference gap bridging via prompt expansion.
    \item The BridgeDiT architecture with DCA fusion, identified through systematic comparison as the optimal interaction strategy for the dual-tower T2SV paradigm.
    \item Comprehensive experiments and analyses across three benchmarks that achieve state-of-the-art results and offer key insights into caption design and fusion architecture choices.
\end{itemize}

\section{Related Works}

\paragraph{Single-Modal Generation.} Recent years have seen significant progress in text-conditional single-modal generation, including Text-to-Video (T2V)~\citep{wang2023modelscope,opensora,lin2024opensora-plan,kong2024hunyuanvideo,kling,sora2024,wan2025wan} and Text-to-Audio (T2A)~\citep{liu2023audioldm,liu2024audioldm,evans2025stableaudioopen,huang2023makeanaudio}. Both domains have evolved from UNets~\citep{3dunet} to Diffusion Transformers (DiT)~\citep{Peebles2022DiT} and adopted efficient paradigms like flow matching~\citep{lipman2022flow}. While these models produce high-quality individual outputs, generating video and audio independently often results in poor semantic and temporal synchronization. Cascaded pipelines address this by conditioning one modality on the other, such as Video-to-Audio (V2A)~\citep{wang2024tiva,cheng2025lova,xing2024seeing,cheng2025mmaudio,wang2025klingaudio,vssflow} or Audio-to-Video (A2V)~\citep{lee2022landscape,jeong2023tpos,yariv2024avalign}. However, these sequential methods suffer from error accumulation~\citep{liu2024syncflow,liu2025javisdit}: artifacts from the first stage inevitably propagate and degrade the final output~\citep{Guan_2025_ICCV,chen2025detectingmitigatinginsertionhallucination}, motivating the shift toward joint audio-video generation.

\paragraph{Audio-Video Joint Generation.}
Existing methods follow two paradigms. The single-tower approach learns the joint audio-video distribution from scratch~\citep{ruan2022mmdiffusion,tang2023codi,sun2024mmldm,wang2024avdit,zhao2025uniform}, but requires vast paired datasets and has shown limited practical success. The dual-tower paradigm has thus emerged as a more practical alternative, leveraging pretrained T2V and T2A backbones and focusing training on a lightweight interaction module for cross-modal fusion. The design of this module is critical. Current strategies include full attention for direct fusion~\citep{wang2025jointdit}, ControlNet-style~\citep{zhang2023controlnet} unidirectional conditioning (video$\rightarrow$audio~\citep{liu2024syncflow} or audio$\rightarrow$video~\citep{weng2025mtv}), and specialized components such as the Prior Estimator in JavisDiT~\citep{liu2025javisdit}. Our work adopts the dual-tower paradigm and contributes a systematic comparison of fusion strategies, identifying Dual Cross-Attention as the optimal mechanism for bidirectional audio-video interaction.

\paragraph{Text Condition in Audio-Video Joint Generation.}
Text prompts serve as the fundamental semantic anchor for synchronizing audio and video modalities. Current strategies fall into two categories. \textit{Shared prompting} methods such as CoDi~\citep{tang2023codi} employ a single, usually visual-centric description to condition both modalities, inherently lacking auditory details. \textit{Unified prompting} methods concatenate visual and auditory descriptions into one input. However, pretrained video and audio backbones are optimized for modality-specific text; feeding a fused prompt creates semantic interference, where visual attributes (\eg, color) act as noise for the audio model and auditory descriptors (\eg, timbre) distract the video model. Recent works LTX-2~\citep{hacohen2026ltx} and Ovi~\citep{low2025ovi} adopt this unified strategy but primarily target speech and dialogue, where lip-speech coupling naturally favors joint text representations. In contrast, our work focuses on foley and sound-effect generation, where the interference problem is more pronounced. We propose a \textit{disentangled prompting} paradigm that generates modality-pure text conditions ($T_V$ and $T_A$) to ensure precise alignment without interference.

\section{Method}
\subsection{Preliminary}
\paragraph{Denoising-based Generative Models} Denoising generative models learn a complex data distribution $p(\mathbf{x})$ by reversing a process that destroys data to a simple Gaussian prior $\mathcal{N}(\mathbf{0}, \mathbf{I})$. Diffusion models~\citep{ho2020denoisingdiffusionprobabilisticmodels} approach this by training a network $\boldsymbol{\epsilon}_\theta$ to predict the noise $\epsilon$ added to a data sample $\mathbf{x}_0$ at timestep $t$: 
\begin{equation}
\mathcal{L}_{\text{DDPM}}(\theta) = \mathbb{E}_{t, \mathbf{x}_0, \boldsymbol{\epsilon}} \left[ || \boldsymbol{\epsilon} - \boldsymbol{\epsilon}_\theta(\sqrt{\bar{\alpha}_t}\mathbf{x}_0 + \sqrt{1 - \bar{\alpha}_t}\boldsymbol{\epsilon}, t) ||^2 \right].
\end{equation}
Flow Matching (FM)~\citep{lipman2022flow} models learn a velocity field $v_\theta$ that transports a noise sample $\mathbf{x}_0$ to a data sample $\mathbf{x}_1$ by approximating the target field $\mathbf{x}_1 - \mathbf{x}_0$. The training objective is:
\begin{equation}
\mathcal{L}_{\text{FM}}(\theta) = \mathbb{E}_{t, \mathbf{x}_0, \mathbf{x}_1} \left[ || v_\theta(t\mathbf{x}_0 + (1-t)\mathbf{x}_1, t) - (\mathbf{x}_1 - \mathbf{x}_0) ||^2 \right].
\end{equation}
Generation in both cases involves starting with a sampled noise and applying the learned network to iteratively denoise and obtain a clean data sample. More background is in Appendix~\ref{app:back}.
\paragraph{Problem Formulation}
For the T2SV task, we adopt the dual-tower paradigm. This approach is highly practical as it leverages the capabilities of pre-trained unimodal models, a video tower $\mathcal{G}_{\theta}^{{V}}$ and an audio tower $\mathcal{G}_{\theta}^{{A}}$. In this setup, the towers independently process their respective text captions, $T_V$ and $T_A$, audio timestep $t_A$ and video timestep $t_V$, noisy audio latent $\mathbf{x}_A(t_A)$ and noisy video latent $\mathbf{x}_V(t_V)$ while a trainable interaction module, $\mathcal{B}_{\theta}^{{AV}}$, facilitates cross-modal communication:
\begin{equation}
(\hat{\mathbf{a}}, \hat{\mathbf{v}}) = \mathcal{G}_{\text{model}}(T_A, T_V,\mathbf{x}_A(t_A), \mathbf{x}_V(t_V), t_A, t_V), ~~
\mathcal{G}_{\text{model}} = \{\mathcal{G}_{\theta}^{{A}}, \mathcal{G}_{\theta}^{{V}}, \mathcal{B}_\theta^{{AV}}\}.
\end{equation}
The final output consists of the predicted audio $\hat{\mathbf{a}}$ and video $\hat{\mathbf{v}}$ noise vector.

\paragraph{Training Objective}
We sum up the loss from the two towers:
\begin{equation*} \label{eq:total_loss_simplified}
    \mathcal{L} = \mathcal{L}_{\text{audio}} + \mathcal{L}_{\text{video}}.
\end{equation*}
The audio tower follows a diffusion training setup using a v-prediction diffusion~\citep{salimans2022vpredict} loss objective. Given the continuous timestep $t_A \in [0, 1]$, the signal and noise scaling factors are $\alpha(t_A)=\cos(t_A\pi/2)$ and $\sigma(t_A)=\sin(t_A\pi/2)$. We denote $\mathbf{x}_A$ as the audio latent vector from the audio Variational AutoEncoder (VAE)~\citep{kingma2022vae} encoder. 
It predicts the target $\alpha(t_A)\boldsymbol{\epsilon}_A - \sigma(t_A)\mathbf{x}_A$ and for the noisy audio latent $\mathbf{x}_A(t_A) = \alpha(t_A)\mathbf{x}_A + \sigma(t_A)\boldsymbol{\epsilon}_A$:
\begin{equation*} \label{eq:audio_loss_simplified}
    \mathcal{L}_{\text{audio}} = \left\| \hat{\mathbf{a}} - (\alpha(t_A)\boldsymbol{\epsilon}_A - \sigma(t_A)\mathbf{x}_A) \right\|^2.
\end{equation*}
The video tower follows a flow matching~\citep{lipman2022flow} loss objective. The corresponding video timestep $t_V$ is defined as $t_V = 1000 \cdot t_A$. $\mathbf{x}_V$ is the video latent vector. 
It predicts the target vector field $\boldsymbol{\epsilon}_V - \mathbf{x}_V$ and for the noisy video latent $\mathbf{x}_V(t_V) = (1 - t_V/1000)\mathbf{x}_V + (t_V/1000)\boldsymbol{\epsilon}_V$:
\begin{equation*} \label{eq:video_loss_simplified}
    \mathcal{L}_{\text{video}} = \left\| \hat{\mathbf{v}} - (\boldsymbol{\epsilon}_V - \mathbf{x}_V) \right\|^2.
\end{equation*}
Here, $\boldsymbol{\epsilon}_A$ and $\boldsymbol{\epsilon}_V$ are Gaussian noise vectors sampled from $\mathcal{N}(\mathbf{0}, \mathbf{I})$. The detailed inference process is further shown in Appendix~\ref{app:infer}.
\begin{figure}[!t]
    \centering
\includegraphics[width=\linewidth]{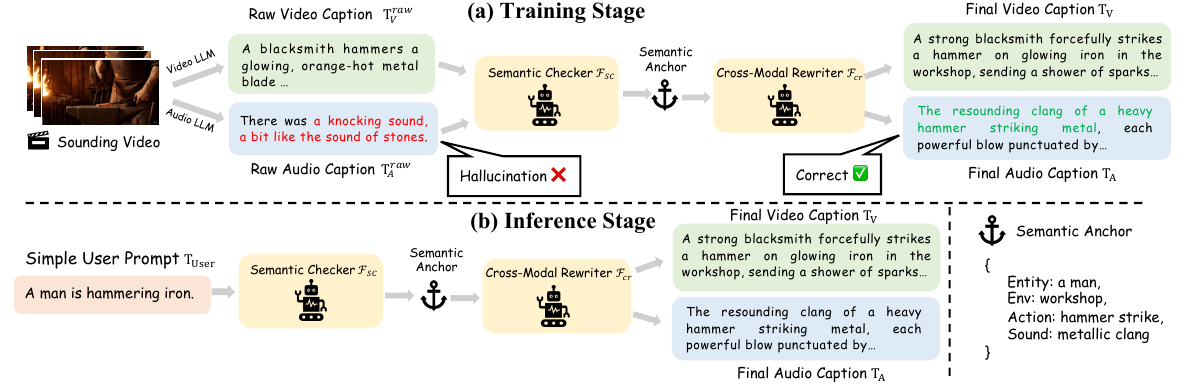}
\vspace*{-0.5cm}
    \caption{Overview of the CRR caption framework. (a) Training: the Semantic Checker $\mathcal{F}_{sc}$ cross-references raw captions from Video and Audio LLMs to filter hallucinations (marked with red) and extract grounded Semantic Anchors $\mathcal{A}$ (structure shown on the right); the Rewriter $\mathcal{F}_{cr}$ then generates modality-pure captions ($T_V$, $T_A$). (b) Inference: the same pipeline expands a concise user prompt into dense, decoupled captions aligned with the training distribution.}
    \vspace*{-0.6cm} 
\label{fig:captioning_framework}
\end{figure}
\subsection{Cross-Referential Rewriter Caption Framework}
\label{sec:crr}
In the dual-tower paradigm, each tower requires its own text condition. Prior methods feed a shared caption to both towers~\cite{liu2024syncflow,tang2023codi,liu2025javisdit,zhao2025uniform}, causing semantic interference. A straightforward fix is to use an LLM to generate separate captions, but without cross-modal verification, audio LLMs frequently hallucinate sounds that have no visual source. The core issue is not caption generation itself, but the lack of a grounding mechanism between the two modal captions.

We address this with the \textbf{Cross-Referential Rewriter (CRR)} caption framework, which decomposes caption generation into two agents connected by a structured bottleneck. First, a Semantic Checker($\mathcal{F}_{sc}$) distills the input into \textit{Semantic Anchors} $\mathcal{A}$---a structured set of verified attributes including core entities, environments, visual actions, and their corresponding acoustic events (\eg, \texttt{\{Entity: worker, Env: workshop, Action: hammer strike, Sound:\\ metallic clang\}}). Then, a Cross-Modal Rewriter ($\mathcal{F}_{cr}$) generates dense, disentangled captions strictly from $\mathcal{A}$:
\begin{equation}
    \mathcal{A} = \mathcal{F}_{sc}(\mathcal{I}), \quad 
    T_V, T_A = \mathcal{F}_{cr}(\mathcal{A}).
\end{equation}
Since $\mathcal{A}$ retains only grounded semantics and discards modality-specific noise, the Rewriter cannot introduce unverified events. This enforces modality purity by construction, not by relying on LLM instruction-following alone. The pipeline is uniform across training and inference---only the input $\mathcal{I}$ changes.

\paragraph{Training Stage:}
Given a sounding video, we first obtain independent raw descriptions from a Video LLM (Qwen2.5-VL~\citep{Qwen2.5-VL}) and an Audio LLM (Qwen2-Audio~\citep{Qwen2-Audio}). While the Video LLM typically produces reliable visual descriptions, the Audio LLM is prone to hallucination---it often generates vague or incorrect sound descriptions due to the inherent ambiguity of audio signals. For example, given a worker hammering scene, the Audio LLM may output ``there was a knocking sound, a bit like the sound of stones'', failing to identify the actual sound source and introducing misleading acoustic content. Directly using such raw captions as training conditions would propagate these errors into the generative model. The Semantic Checker addresses this by cross-referencing both raw captions ($\mathcal{I} = (T_V^{raw}, T_A^{raw})$): it uses the visual description as a grounding reference to verify, correct, and filter the audio description, extracting only semantically consistent events into $\mathcal{A}$. The Rewriter then expands $\mathcal{A}$ into dense, modality-pure captions ($T_V$, $T_A$) that are both semantically aligned and free of cross-modal contamination.

\paragraph{Inference Stage:}
Users typically provide concise prompts (\eg, ``a worker is hammering iron'') that are far shorter and less detailed than the dense training captions, creating a significant distribution gap. If fed directly to the model, such brief inputs lead to degraded generation quality (validated in Table~\ref{tab:caption_ablation}). Here $\mathcal{I} = T_{user}$, and the Semantic Checker applies the same verification logic but switches from cross-referencing to context inference: it deduces the implicit visual scenes and acoustic events entailed by the brief prompt (\eg, ``a worker striking iron'' $\rightarrow$ $\mathcal{A}$=\texttt{\{Entity: worker, Env: workshop, Action: hammer strike, Sound: \\metallic clang\}}). The Rewriter then expands these anchors into detailed, disentangled $T_V$ and $T_A$ that match the density of training captions, bridging the distribution gap without requiring complex prompt engineering from users. Detailed agent instructions and system prompts for the CRR framework are provided in Appendix~\ref{app:prompt}.
\begin{figure}[!t]
    \centering
    \vspace*{-0.5cm}
    \includegraphics[width=\linewidth]{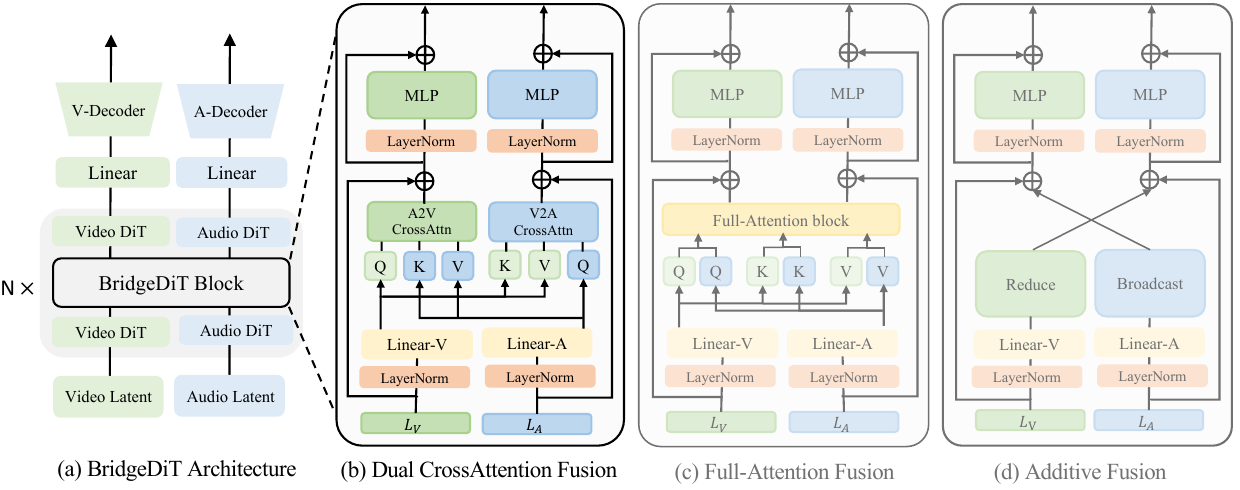}
    \caption{\textbf{Overview of the BridgeDiT Architecture}. (a): The overall dual-tower architecture. Parallel video and audio DiT streams are connected by our proposed BridgeDiT Block at specific layers.  \textit{Right}: Details
    of fusion strategies within the block, showcasing our proposed Dual Cross-Attention (DCA) (b) alongside the Full-Attention (c) and Additive Fusion (d) baselines. The Unidirectional Cross-Attention baseline (omitted for brevity) shares the same structure as DCA but utilizes only a single direction (V2A or A2V) of the information flow.}
    \vspace*{-0.5cm}
    \label{fig:archi}
\end{figure}

\subsection{Audio-Video Bridge Interaction Mechanism}

A key challenge in the dual-tower paradigm is how to exchange information between the video and audio towers to achieve synchronization. This is non-trivial in our setting: both towers are pretrained on separate modalities and remain largely frozen, with features lying on two disjoint manifolds $\mathcal{M}_V$ and $\mathcal{M}_A$. This makes the fusion problem fundamentally different from models trained from scratch (\eg, MMDiT~\citep{esser2024mmdit}), where all modalities converge into a shared space.

\paragraph{BridgeDiT Architecture.}
As shown in Figure~\ref{fig:archi}~(a), BridgeDiT inserts $N$ lightweight interaction blocks between the parallel video and audio DiT streams. Each block performs cross-modal information exchange and returns updated features to both towers. We place the blocks at early-to-mid layers, where features balance spatial-temporal detail and semantic abstraction (see Appendix~\ref{app:placement}).

\paragraph{Dual Cross-Attention Fusion (Ours).}
As shown in Figure~\ref{fig:archi}~(b), our Dual Cross-Attention (DCA) mechanism updates the video latent $L_V$ and audio latent $L_A$ through two symmetric cross-attention streams. Taking the Audio-to-Video (A2V) stream as an example, the video latent forms the query while the audio latent provides key and value:

\begin{gather}
    Q_V \!=\! \text{Linear}_{Q_V}(\text{LN}(L_V)),~~ K_A \!=\! \text{Linear}_{K_A}(\text{LN}(L_A)),~~ V_A \!=\! \text{Linear}_{V_A}(\text{LN}(L_A)), \\[2pt]
    L'_V = \text{Attention}(Q_V, K_A, V_A) + L_V.
\end{gather}
The V2A stream is perfectly symmetric: the audio latent serves as query, while the video latent provides key and value, yielding $L'_A$. While cross-attention is a well-established operation, we argue that DCA is the right choice for the pretrain backbone-based dual-tower setting for two reasons: (1)~DCA preserves each tower's feature space---unlike Full-Attention, which concatenates both modalities and mixes intra- and inter-modal dependencies in one attention matrix, DCA keeps streams separate so each tower only receives targeted cross-modal cues; (2)~this separation yields an easier optimization target---rather than merging $\mathcal{M}_V$ and $\mathcal{M}_A$ into a shared space with limited trainable parameters, DCA learns a lightweight bridge ($\mathcal{M}_V \!\leftrightarrow\! \mathcal{M}_A$) between two established manifolds. We validate this analysis in Section~\ref{sec:fusion_ablation}.

To justify our design, we compare Dual Cross-Attention Fusion against three baselines under identical conditions (Figure~\ref{fig:archi}~(c,d)):

\noindent\textit{Full-Attention Fusion}~\citep{wang2025jointdit} concatenates both latents for joint self-attention, allowing all-to-all interaction but forcing a shared space:

\begin{equation}
    L'_{\text{cat}} = \text{Attention}(\text{Cat}(Q_V, Q_A),\; \text{Cat}(K_V, K_A),\; \text{Cat}(V_V, V_A)) + \text{Cat}(L_V, L_A).
\end{equation}

\noindent\textit{Additive Fusion}~\citep{ishii2024ssvg} projects and adds cross-modal features to the residual stream---lightweight but non-selective:

\begin{equation}
    L'_V = L_V + \text{Linear}_{A \to V}(\text{LN}(L_A)), \quad L'_A = L_A + \text{Linear}_{V \to A}(\text{LN}(L_V)).
\end{equation}

\noindent\textit{Unidirectional Cross-Attention}~\citep{liu2024syncflow,weng2025mtv} conditions one modality on the other in a ControlNet~\citep{zhang2023controlnet} style, restricting flow to one direction (V2A or A2V). Its formulation is identical to one half of DCA; we implement both variants.

\section{Experiments}

\subsection{Experimental Setup}
\paragraph{Implementation Details}
For the video backbone, we use Wan 2.1 (1.3B)~\citep{wan2025wan} with a UMT5-XXL~\citep{chung2023unimax} text encoder, generating 81 frames at 15fps and 480p resolution. For the audio backbone, we employ Stable Diffusion Audio Open 1.0~\citep{evans2025stableaudioopen} (SDA for short) with a T5-base text encoder~\citep{2020t5}, generating audio at 44.1kHz. The total generation length is 5.4 seconds. Our BridgeDiT architecture consists of 4 BridgeDiT Blocks uniformly inserted between the layers of both towers. For the CRR caption framework, both the Semantic Checker and Cross-Modal Rewriter are implemented using Qwen2.5-72B~\citep{qwen2.5}; the raw captions are generated by Qwen2.5-VL-72B~\citep{Qwen2.5-VL} (video) and Qwen2-Audio~\citep{Qwen2-Audio} (audio). We train our models separately for each dataset. More details are in Appendix~\ref{app:exp_setup}.

\paragraph{Datasets.}
We evaluate on three datasets: (1)~AVSync15~\citep{linz2024avsync}, a subset of VGGSound~\citep{Chen20vggsound} containing synchronized audio-video pairs across 15 categories, split into 1350 training and 150 test videos; (2)~VGGSound-SS~\citep{Chen21vggss}, a sound source localization dataset from VGGSound~\citep{Chen20vggsound} with 5,158 videos across 220 classes, from which we randomly sample 150 for testing; and (3)~Landscape~\citep{lee2022landscape}, comprising 928 videos across 9 scenic categories. Since these datasets lack standard captions, we generate them using the CRR framework. All videos are standardized to 5.4 seconds via random cropping or padding.

\paragraph{Baselines.}
We compare against baselines spanning five T2SV generation paradigms. Since our work focuses on foley and sound-effect generation, we exclude speech-focused methods such as LTX-2~\citep{hacohen2026ltx} and Ovi~\citep{low2025ovi}.
We distinguish two groups for fairness. \textit{Retrained baselines} (Wan+SDA, SSVG~\citep{ishii2024ssvg}, MTV~\citep{weng2025mtv}) are trained on each benchmark using our CRR captions, sharing identical data and text conditions with our method. \textit{Pre-trained baselines} (MMAudio~\citep{cheng2025mmaudio}, SeeingHearing~\citep{xing2024seeing},ThinkSound~\cite{liu2025thinksoundchainofthoughtreasoningmultimodal},PrismAudio~\cite{liu2025prismaudiodecomposedchainofthoughtsmultidimensional}, JavisDiT~\citep{liu2025javisdit}, JointDiT~\citep{wang2025jointdit}, CoDi~\citep{tang2023codi}) are evaluated with released checkpoints trained on substantially larger datasets; we feed them prompts in their original expected format to ensure optimal performance.
The five paradigms are:
(1)~T$\rightarrow$V$\|$T$\rightarrow$A: Independent generation with interaction modules disabled (Wan+SDA).
(2)~T$\rightarrow$V$\rightarrow$A: Wan-1.3B generates video, then MMAudio~\citep{cheng2025mmaudio} or SeeingHearing~\citep{xing2024seeing} synthesizes audio.
(3)~T$\rightarrow$A$\rightarrow$V: SDA~\citep{evans2025stableaudioopen} generates audio, then TPos~\citep{jeong2023tpos} or TempoToken~\citep{yariv2024avalign} produces video.
(4)~T$\rightarrow$I$\rightarrow$AV: Qwen-Image~\citep{wu2025qwenimagetechnicalreport} generates an image for JointDiT~\citep{wang2025jointdit} to jointly produce video and audio.
(5)~T$\rightarrow$AV: Joint models including JavisDiT~\citep{liu2025javisdit}, SSVG~\citep{ishii2024ssvg}, CoDi~\citep{tang2023codi}, and MTV~\citep{weng2025mtv}.
Details are in Appendix~\ref{app:baselines}.
\newcommand{\std}[1]{\,$\pm$\,#1}
\definecolor{mypalelue}{rgb}{0.85, 0.92, 1.0}

\setlength{\textfloatsep}{10pt}      

\begin{table}[t]
\centering
\caption{Automatic evaluation on the AVSync15 dataset. \textbf{Best} and \underline{second-best} are highlighted.}
\vspace*{-8pt}
\large
\renewcommand{\arraystretch}{1.2}
\resizebox{\textwidth}{!}{%
\begin{tabular}{ll|cccc|cc|cc}

\toprule[1.5pt]
\raisebox{-1.5ex}{\textbf{Process}} & \raisebox{-1.5ex}{\textbf{Method}}  & \multicolumn{4}{c|}{\textbf{Quality}}  & \multicolumn{2}{c|}{\textbf{Text Alignment}} & \multicolumn{2}{c}{\textbf{Synchronization}} \\
& & FVD$\downarrow$ & KVD$\downarrow$ & FAD$\downarrow$ & KL$\downarrow$ & CLIPSIM$\uparrow$ & CLAP$\uparrow$ & VA-IB$\uparrow$ & AV-Align$\uparrow$ \\
\hline
T $\rightarrow$ V $\|$ T $\rightarrow$ A & Wan + SDA & \underline{828.33} & \underline{22.56} & 11.90 & 3.17 & 28.12 & 30.78 & 26.22 & 0.205 \\
\hline
\multirow{2}{*}{T $\rightarrow$ V + V $\rightarrow$ A} 
& Wan + MMAudio & \underline{828.33} & \underline{22.56} & 7.98 & \underline{1.40} & 28.12 & \underline{34.64}	& 33.50	& \underline{0.243} \\
& Wan + SeeingHearing & \underline{828.33} & \underline{22.56} & 14.21 & 2.89 &28.12	& 25.52 & \textbf{35.87} & 0.208 \\
& Wan + ThinkSound & \underline{828.33} & \underline{22.56} & 17.31 & 5.58 & 28.12 & 30.92 & 20.70 & 0.169 \\
& Wan + PrismAudio & \underline{828.33} & \underline{22.56} & 9.79 & 2.34 & 28.12 & 33.53 & 25.25 & 0.233 \\
\hline
\multirow{2}{*}{T $\rightarrow$ A + A $\rightarrow$ V} 
& SDA +  TPos & 1975.22 & 92.95 & 11.90	& 3.17  & 18.64 & 30.78&  25.45 & 0.176  \\
& SDA +  TempoToken & 1516.53	& 42.30	& 11.90	& 3.17 & 20.56 &	30.78 & 17.63 & 0.215 \\
\hline
T $\rightarrow$ I + I $\rightarrow$ VA
& JointDiT & 992.71 & 25.20 & \underline{6.51} & 1.77 &\textbf{29.94} & 30.34 & 34.17 &  0.156 \\
\hline
\multirow{5}{*}{T $\rightarrow$ VA} 
& JavisDiT & 878.70 & 23.23 & 13.48 &3.50 &	28.05 &22.99 &20.75 &	0.158 \\
& SSVG & 1028.78  & 31.44  & 14.35 & 4.35  & 25.78 & 23.67 & 22.45 & 0.126  \\
& MTV & 982.09 & 24.60 & 16.46 & 3.53 & 27.66 & 21.22 &	15.84 &	0.149 \\
& CoDi & 1387.14 & 39.24 & 16.56 & 5.24 & 24.96 & 17.94	& 10.79 &  0.081\\
\rowcolor{mypalelue} 
\cellcolor{white} & \textbf{BridgeDiT (ours)} & \textbf{765.74} & \textbf{21.33} & \textbf{5.34} & \textbf{1.30} & \underline{28.52} & \textbf{35.95} & \underline{34.59} & \textbf{0.275} \\
\bottomrule
\end{tabular}%
\label{tab:avsync}
}
\end{table}

\paragraph{Automatic Evaluation Metrics}
We evaluate the T2SV task from three different perspectives: generation quality, text alignment, and audio-video synchronization. (1) Generation Quality. For video quality, we employ the Fréchet Video Distance (FVD)~\citep{unterthiner2018fvd} and Kernel Video Distance (KVD)~\citep{unterthiner2018fvd} with the Inflated 3D classifier~\cite{carreira2018i3d}. For audio quality, we use the Fréchet Audio Distance~\citep{kilgour2018fad} (FAD) and the Kullback-Leibler~\citep{wang2024tiva}  (KL) divergence score. (2) Text Alignment. We evaluate video and audio text alignment separately. We use CLIPSIM~\citep{radford2021clip}score to evaluate video-text alignment and CLAP~\citep{elizalde2023clap} score to measure audio-text alignment. (3) Audio-Video Synchronization. We evaluate both semantic audio-video synchronization using the ImageBind score (VA-IB)~\citep{girdhar2023imagebind} and temporal synchronization using the AV-Align score~\citep{yariv2024avalign}. Recognizing the limitations of automatic metrics, we supplement these with human evaluations in Section~\ref{user_study}.


\definecolor{mypalelue}{rgb}{0.85, 0.92, 1.0}
\begin{wrapfigure}{R}{0.6\textwidth}
\centering\small
\vspace{-0.45in}
\captionof{table}{
Performance on VGGSound-SS and Landscape. AV denotes
AV-Align metric here. \textbf{Best} and \underline{second-best} are highlighted.
}
\resizebox{0.6\textwidth}{!}{
\begin{tabular}{l|c c c|c c c}
    \toprule
     \textbf{Method} & \multicolumn{3}{c}{\textbf{VGGSound-SS}}  & \multicolumn{3}{c}{\textbf{LandScape}} \\
     & FVD$\downarrow$ & FAD$\downarrow$ & AV-Align$\uparrow$ & FVD$\downarrow$ & FAD$\downarrow$ & AV-Align$\uparrow$\\
     \midrule
\rowcolor{gray!20} 
     \multicolumn{7}{l}{\textit{T $\rightarrow$ V $\|$ T $\rightarrow$ A}} \\ 
     Wan + SDA & 737.96 & 8.05 & 0.238 & 700.18 & 5.80 & 0.177 \\
    \midrule
\rowcolor{gray!20} 
     \multicolumn{7}{l}{\textit{T $\rightarrow$ V + V $\rightarrow$ A}} \\
     Wan + MMAudio & 737.96 & \textbf{5.39} & \underline{0.333} & 700.18 & 
     \underline{5.31} & \underline{0.218}\\
     Wan + SeeingHearing & 737.96 & 8.02 & 0.321 & 700.18 & 7.59 & 0.166\\
\midrule
\rowcolor{gray!20} 
     \multicolumn{7}{l}{\textit{T $\rightarrow$ A + A $\rightarrow$ V}} \\
     SDA + TPos & 1732.47 & 8.05 & 0.163 & 1837.42 & 5.80 & 0.143 \\
     SDA + TempoToken & 1942.94 & 8.05 & 0.242 & 2089.05 & 5.80 & 0.206 \\
\midrule
\rowcolor{gray!20} 
     \multicolumn{7}{l}{\textit{T $\rightarrow$ I + I $\rightarrow$ VA}} \\
     JointDiT & 866.59 & 7.19 & 0.125 & 937.09 & 6.35 & 0.075\\
\midrule
\rowcolor{gray!20} 
     \multicolumn{7}{l}{\textit{T $\rightarrow$ VA}} \\
     JavisDiT & \underline{637.50} & 7.78 & 0.179 & \underline{668.87} & 9.22 & 0.185 \\
     SSVG & 1032.87 & 8.86 & 0.148 & 1186.39 & 8.97 & 0.143\\
     CoDi & 1203.68 & 8.38 & 0.113 & 1220.31 & 13.75 & 0.082\\
\midrule
\rowcolor{mypalelue} 
     \textbf{BridgeDiT (ours)} & \textbf{615.28} & \underline{6.01} & \textbf{0.362} & \textbf{628.07} & \textbf{4.78} & \textbf{0.258}\\
     \bottomrule
\end{tabular}
}
\label{tab:vgg-landscape}
\vspace{-0.3in}
\end{wrapfigure}
\subsection{Main Results: Comparison with Baselines}

We present results on three datasets in Table~\ref{tab:avsync} and Table~\ref{tab:vgg-landscape}. Our BridgeDiT surpasses all baselines on most metrics, including video quality (FVD, KVD), audio quality (FAD, KL), audio-text alignment (CLAP), and temporal synchronization (AV-Align).
Among the retrained baselines that share identical CRR captions and training data, BridgeDiT consistently outperforms SSVG and MTV, directly validating the architectural advantage of our DCA fusion mechanism. Compared with Wan+SDA---which is equivalent to our architecture with interaction modules removed---BridgeDiT shows clear improvements across all metrics, confirming that cross-modal interaction is essential for both generation quality and synchronization. BridgeDiT also outperforms all pipelined baselines, demonstrating that joint generation effectively mitigates the error accumulation inherent to sequential approaches.
We note two minor exceptions. Our CLIPSIM (28.52) is slightly below JointDiT~\citep{wang2025jointdit} (29.94), which we attribute to its stronger T2I backbone Qwen-Image~\citep{wu2025qwenimagetechnicalreport}. Our VA-IB (34.59) is also surpassed by SeeingHearing (35.87), which directly uses ImageBind~\citep{girdhar2023imagebind} as classifier guidance during inference---optimizing for this exact metric. As shown in Table~\ref{tab:vgg-landscape}, BridgeDiT achieves competitive results on VGGSound-SS and Landscape as well, confirming generalization across domains.
However, automatic metrics do not always reflect perceptual quality---for instance, a method optimizing for a specific metric (\eg, SeeingHearing for VA-IB) may score high without producing genuinely better outputs. To obtain a more reliable assessment, we conduct a comprehensive user study in Section~\ref{user_study}.

\definecolor{mypalelue}{rgb}{0.85, 0.92, 1.0}
\begin{wrapfigure}{R}{0.5\textwidth}
\centering\small
\vspace{-0.18in}
\captionof{table}{
User study on the AVSync15 test set with 20 raters}

\resizebox{0.5\textwidth}{!}{
\begin{tabular}{l|c c c c c}
\toprule
\textbf{Method} & VQ $\uparrow$ & AQ$\uparrow$ & TA$\uparrow$ & Sync$\uparrow$ & Overall$\uparrow$ \\
\midrule
Wan+SDA & 3.16 & 2.93 & 2.75 & 2.47 & 2.79 \\
Wan+Seeing & 3.16 & 2.83 & 2.91 & 2.54 & 2.85 \\
Wan+MMAudio & 3.16 & 3.18 & 3.07 & 2.77 & 3.06 \\
SDA+TmToken & 1.97 & 2.36 & 2.00 & 1.83 & 2.04 \\
JointDiT & 2.74 & 2.81 & 2.79 & 2.45 & 2.69 \\
JavisDiT & 2.47 & 2.39 & 2.36 & 2.14 & 2.36 \\
\rowcolor{mypalelue} 
\textbf{BridgeDiT (ours)} & \textbf{3.40} & \textbf{3.33} & \textbf{3.33} & \textbf{3.09} & \textbf{3.34} \\
\bottomrule
\end{tabular}
}
\label{tab:user_study}
\vspace{-0.3in}
\end{wrapfigure}
\subsection{User Study}
\label{user_study}
To further validate our model with human preference, we conduct a user study on the AVSync15 test set with 150 samples. 20 evaluators rate the generated sounding videos on a 0--5 scale with 0.5-point increments across five criteria: Video Quality (VQ), Audio Quality (AQ), Text Alignment (TA), Synchronization (Sync), and Overall quality. As shown in Table~\ref{tab:user_study}, our BridgeDiT model is rated highest across all five dimensions, significantly outperforming all baselines, with the pipelined Wan+MMAudio ranking second. Notably, some baselines that achieve strong scores on specific automatic metrics (as in Table~\ref{tab:avsync}) are not preferred by human evaluators, reinforcing our observation that automatic metrics do not fully capture perceptual quality. This further highlights the importance of human evaluation as a complementary assessment for the T2SV task. Detailed annotation guidelines are provided in Appendix~\ref{app:human}.

\subsection{Ablation Study on CRR Caption Framework}

\begin{table}[t]
\centering
\caption{Ablation study of the CRR caption framework on AVSync15.}
\vspace*{-8pt}
\small
\renewcommand{\arraystretch}{1.2}
\resizebox{\textwidth}{!}{%
\begin{tabular}{l|cccccc}
\toprule[1.5pt]
\textbf{Caption Strategy} & \textbf{FVD}~$\downarrow$ & \textbf{FAD}~$\downarrow$ & \textbf{CLIPSIM}~$\uparrow$ & \textbf{CLAP}~$\uparrow$ & \textbf{VA-IB}~$\uparrow$ & \textbf{AV-Align}~$\uparrow$ \\
\hline
\rowcolor{gray!15}
\multicolumn{7}{l}{\textit{Shared text condition ($T_V = T_A$)}} \\
Video Caption ($T_V$) &  788.65 & 16.46 & 28.34 & 9.67 & 18.54 & 0.176\\
Audio-Video Caption ($T_{AV}$) & 1362.83 & 13.75 & 25.81 & 26.37 & 26.82 & 0.185 \\
\hline
\rowcolor{gray!15}
\multicolumn{7}{l}{\textit{Disentangled text condition ($T_V \neq T_A$)}} \\
Raw LLM Captions (w/o CRR) &  924.36 & 19.42 & 26.22 & 8.43 & 17.84 & 0.161\\
Direct Rewrite (w/o Semantic Anchors) & 787.32 & 15.74 & 28.34 & 27.34  & 28.82 & 0.224 \\
\rowcolor{mypalelue}
\textbf{CRR (ours)} & \textbf{765.74} & \textbf{5.34} & \textbf{28.52} & \textbf{35.95} & \textbf{34.59} & \textbf{0.275}\\
\hline
\rowcolor{gray!15}
\multicolumn{7}{l}{\textit{CRR inference ablation}} \\
w/o Prompt Expansion & 1463.81 & 20.84 & 25.66 & 22.45 & 23.36 & 0.155 \\
\bottomrule
\end{tabular}%
\label{tab:caption_ablation}
}
\end{table}

We ablate the CRR caption framework on AVSync15 by systematically varying the text conditioning strategy. Results are shown in Table~\ref{tab:caption_ablation}. We organize the comparisons into three groups: shared text conditions, disentangled text conditions, and inference-stage ablation.

\paragraph{Shared vs.\ Disentangled Conditioning.}
When both towers receive a shared caption, performance degrades regardless of caption content. Using only the video caption ($T_V$) starves the audio tower of acoustic details, resulting in poor audio-text alignment (CLAP: 9.67). The unified Audio-Video Caption ($T_{AV}$) retains some audio information but introduces semantic interference---visual attributes distract the audio model and vice versa---leading to severely degraded video quality (FVD: 1362.83). These results confirm that disentangled, modality-specific conditioning is essential for the dual-tower paradigm.

\paragraph{The Role of Semantic Anchors.}
Within the disentangled group, we compare three strategies of increasing sophistication. Raw LLM Captions directly use unverified outputs from the Video and Audio LLMs without any rewriting, yielding the worst disentangled performance (FAD: 19.42, AV-Align: 0.161) due to hallucinations and cross-modal conflicts. Direct Rewrite removes the Semantic Anchor bottleneck and lets a single LLM rewrite raw captions into modality-specific pairs in one step. This improves over raw captions, but still lags behind full CRR by a large margin, particularly in audio quality (FAD: 15.74 vs.\ 5.34) and synchronization (AV-Align: 0.224 vs.\ 0.275). The gap demonstrates that the structured Semantic Anchors are not merely a prompt engineering trick---they serve as a critical grounding mechanism that prevents the Rewriter from introducing unverified events, validating our core design argument in Section~\ref{sec:crr}.

\paragraph{Inference-Stage Prompt Expansion.}
Removing prompt expansion at inference (\ie, feeding concise user prompts directly) causes the most severe degradation (FVD: 1463.81, FAD: 20.84). This confirms that the distribution gap between brief user inputs and dense training captions is the single largest bottleneck for generation quality, and that CRR's prompt expansion is essential to bridge it.



\subsection{Ablation Study on Fusion Mechanisms}
\label{sec:fusion_ablation}

\begin{figure}[t]
    \centering
    \begin{minipage}[t]{0.48\textwidth}
        \centering
        \includegraphics[width=\linewidth]{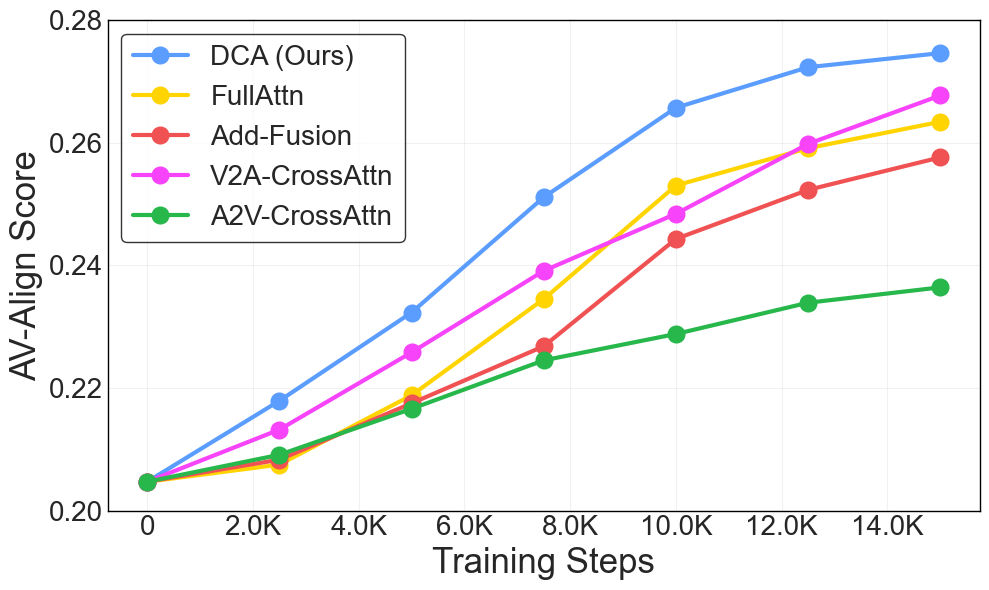}
    \end{minipage}
    \hfill
    \begin{minipage}[t]{0.48\textwidth}
        \centering
        \includegraphics[width=\linewidth]{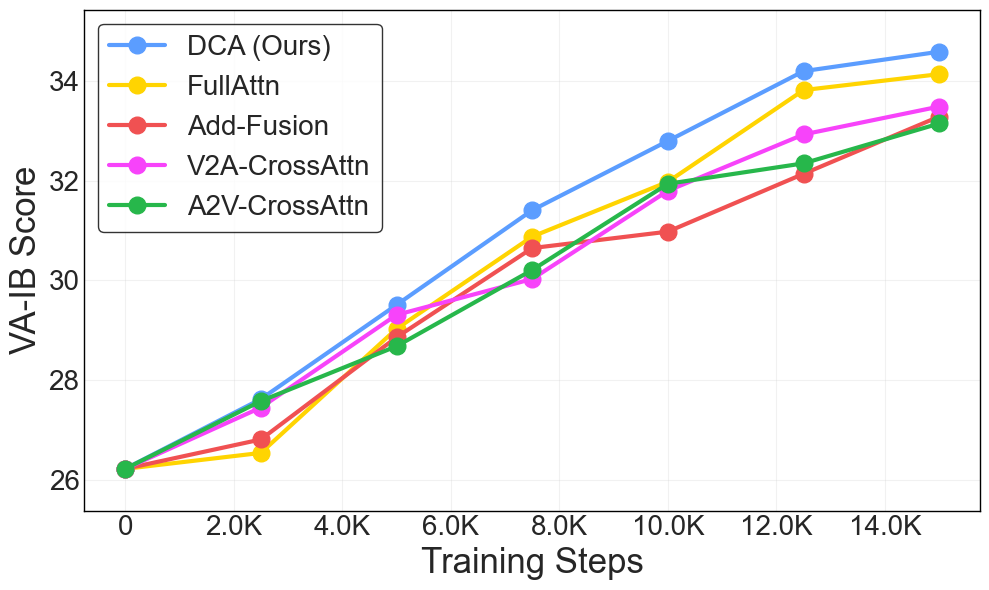}
    \end{minipage}
    \caption{Training dynamics of different fusion mechanisms on AV-Align and VA-IB scores. DCA achieves the highest synchronization throughout training.}
    \label{fig:fusion_ablation}
\end{figure}
\definecolor{mypalelue}{rgb}{0.85, 0.92, 1.0}

\begin{table}[t]
    \centering
    \small
    \caption{\textbf{Comprehensive evaluation on fusion mechanisms.} We compare our DCA mechanism against various baselines on the AVSync15 dataset. Metrics include video quality (FVD), audio quality (FAD), text-alignment (CLIPSIM, CLAP), and audio-visual synchronization (VA-IB, AV-Align). \textbf{Bold} indicates the best performance.}
    \label{tab:fusion_comprehensive}
    \resizebox{\textwidth}{!}{
    \begin{tabular}{l|cc|cc|cc}
        \toprule
        \textbf{Model} & \textbf{FVD} $\downarrow$ & \textbf{FAD} $\downarrow$ & \textbf{CLIPSIM} $\uparrow$ & \textbf{CLAP} $\uparrow$ & \textbf{VA-IB} $\uparrow$ & \textbf{AV-Align} $\uparrow$ \\
        \midrule
        No-Fusion & 828.33 & 11.90 & 28.12 & 30.78 & 26.22 & 0.205 \\
        Full-Fusion & 781.03 & 5.62 & 28.43 & 32.28 & 34.14 & 0.253 \\
        V2A-CrossAttn & 813.02 & 6.21 & 28.24 & 35.85 & 34.20 & 0.268 \\
        A2V-CrossAttn & \textbf{746.37} & 5.91 & 28.49 & 31.25 & 31.54 & 0.236 \\
        Additive-Fusion & 772.23 & 5.72 & 28.20 & 34.77 & 28.34 & 0.258 \\
        \midrule
        \rowcolor{mypalelue} 
        \textbf{DCA (Ours)} & 765.74 & \textbf{5.34} & \textbf{28.52} & \textbf{35.95} & \textbf{34.59} & \textbf{0.275} \\
        \bottomrule
    \end{tabular}
    }
\end{table}

We compare our DCA mechanism against four baselines under identical CRR captions and training conditions: Full-Attention~\citep{wang2025jointdit}, Additive Fusion~\citep{ishii2024ssvg}, V2A-CrossAttn~\citep{liu2024syncflow}, and A2V-CrossAttn~\citep{weng2025mtv}. The comprehensive quantitative results of this evaluation are detailed in Table~\ref{tab:fusion_comprehensive} while the corresponding training dynamics are visualized in Figure~\ref{fig:fusion_ablation}.

\paragraph{Results.}
The No-Fusion baseline performs worst on most metrics, confirming that cross-modal interaction is essential. Among all fusion strategies, DCA achieves the best overall performance: it ranks first on audio quality (FAD: 5.34), text alignment (CLIPSIM: 28.52, CLAP: 35.95), and both synchronization metrics (VA-IB: 34.59, AV-Align: 0.275). The A2V-CrossAttn variant achieves a slightly lower FVD (746.37), but at the cost of significantly weaker synchronization (AV-Align: 0.236), indicating that unidirectional flow improves one modality while neglecting the other. Figure~\ref{fig:fusion_ablation} further shows that DCA consistently leads on both AV-Align and VA-IB throughout training, while other methods plateau or fluctuate at lower levels.

\paragraph{Why DCA Outperforms Full-Attention.}
Full-Attention (FA) is widely used for multimodal fusion in models trained from scratch (\eg, MMDiT~\citep{esser2024mmdit}), yet it underperforms DCA in our setting. We attribute this to the frozen-backbone constraint. In MMDiT, all parameters are trained jointly, allowing features from different modalities to gradually converge into a unified manifold $\mathcal{M}_{joint}$. In our framework, the video and audio backbones are independently pretrained and largely frozen, so their features reside on two disjoint manifolds $\mathcal{M}_V$ and $\mathcal{M}_A$.
FA concatenates both latents into a single sequence and applies global self-attention, forcing intra-modal and inter-modal dependencies into one shared matrix. With limited trainable parameters, this effectively attempts to merge $\mathcal{M}_V$ and $\mathcal{M}_A$ into a shared space---a difficult optimization target that leads to the unstable synchronization curves observed in Figure~\ref{fig:fusion_ablation}. DCA, by contrast, keeps the two streams separate and only exchanges targeted cross-modal cues through dedicated attention heads. It learns a lightweight bridge ($\mathcal{M}_V \leftrightarrow \mathcal{M}_A$) between two established manifolds rather than attempting to unify them. This is a fundamentally easier task given the parameter budget, resulting in more stable training and stronger synchronization.
This analysis also explains why the two unidirectional variants show asymmetric behavior: A2V-CrossAttn improves video quality (FVD: 746.37) by conditioning video on audio, but leaves the audio stream unrefined; V2A-CrossAttn shows the opposite pattern. DCA resolves this by enabling bidirectional exchange, ensuring that both modalities benefit from cross-modal information simultaneously.

\subsection{Case Studies}
Figure~\ref{fig:teaser} presents case studies that highlight the capabilities of our BridgeDiT model. Powered by the combination of our CRR caption framework and the DCA fusion mechanism, our model generates high-quality sound videos that are semantically synchronized, temporally synchronized, and highly aligned with the text conditions. The first case (the blacksmith) showcases precise temporal synchronization, as the visual impact of the hammer striking the iron aligns perfectly with the sharp ``clang" event in the audio spectrogram. The second case (the saxophone player) demonstrates strong text alignment; the generated video accurately depicts key entities from the visual prompt, including the ``saxophone" and ``metal drum", while the audio faithfully synthesizes the complex soundscape described in the audio prompt. Notably, these two cases represent distinct challenges: the blacksmith involves sparse, impulsive sound events requiring precise temporal alignment, whereas the saxophone scene involves continuous, overlapping audio sources demanding accurate semantic disentanglement---our model handles both effectively. We provide more cases in the Appendix~\ref{app:case_study}.

\section{Conclusion}
In this work, we address two fundamental challenges in Text-to-Sounding-Video generation: suboptimal text conditioning and ineffective cross-modal interaction. For text conditioning, we propose the CRR caption framework, a dual-agent pipeline where a Semantic Checker extracts grounded Semantic Anchors and a Cross-Modal Rewriter produces disentangled, modality-pure captions, eliminating modal interference and bridging the training-inference gap. For cross-modal interaction, we propose BridgeDiT, connecting pretrained video and audio backbones via Dual Cross-Attention (DCA), which we show through systematic comparison to be the optimal fusion strategy for the dual-tower paradigm.  Experiments on three benchmarks with human evaluations demonstrate state-of-the-art performance, and ablation studies validate the necessity of each component. We believe these findings provide a solid foundation for scaling T2SV systems to longer durations and more diverse acoustic scenarios.

\paragraph{Limitations.} Our evaluation currently relies on relatively small-scale benchmarks (up to 5K videos), which may not fully capture the diversity of real-world acoustic environments. We also do not yet support speech or music generation: extending to these modalities is hindered by the scarcity of open-source pre-trained models and would require resource-intensive training from scratch. Additionally, exploring post-training refinement techniques, such as Reinforcement Learning from Human Feedback (RLHF), presents a promising pathway to further optimize audio-visual synchronization and align with human perception.

\section*{Acknowledgement}
This work is supported by the National Natural Science
Foundation of China (No. 62276268) and the Outstanding
Innovative Talents Cultivation Funded Programs 2022 of
Renmin University of China. Experiments of this work used
the Bridges2 system at PSC and Delta system at NCSA
through allocations CIS210014 and IRI120008P from the
Advanced Cyberinfrastructure Coordination Ecosystem:
Services \& Support (ACCESS) program.
\makeatletter
\renewcommand{\bibsection}{%
  \section*{References}%
  \@mkboth{\MakeUppercase\refname}{\MakeUppercase\refname}%
}
\makeatother

\setlength{\bibsep}{3pt} 

\bibliographystyle{splncs04}
\bibliography{main}

@misc{kong2024hunyuanvideo,
      title={HunyuanVideo: A Systematic Framework For Large Video Generative Models}, 
      author={Weijie Kong, Qi Tian},
      year={2024},
      archivePrefix={arXiv preprint arXiv:2412.03603},
      primaryClass={cs.CV},
      url={https://arxiv.org/abs/2412.03603}, 
}

@article{opensora,
  title={Open-sora: Democratizing efficient video production for all},
  author={Zheng, Zangwei and Peng, Xiangyu and Yang, Tianji and Shen, Chenhui and Li, Shenggui and Liu, Hongxin and Zhou, Yukun and Li, Tianyi and You, Yang},
  journal={arXiv preprint arXiv:2412.20404},
  year={2024}
}

@article{lin2024opensora-plan,
  title={Open-Sora Plan: Open-Source Large Video Generation Model},
  author={Lin, Bin and Ge, Yunyang and Cheng, Xinhua and Li, Zongjian and Zhu, Bin and Wang, Shaodong and He, Xianyi and Ye, Yang and Yuan, Shenghai and Chen, Liuhan and others},
  journal={arXiv preprint arXiv:2412.00131},
  year={2024}
}

@misc{sora2024,
  title={Video generation models as world simulators},
  author={Tim Brooks and Bill Peebles and Connor Holmes and Will DePue and Yufei Guo and Li Jing and David Schnurr and Joe Taylor and Troy Luhman and Eric Luhman and Clarence Ng and Ricky Wang and Aditya Ramesh},
  year={2024},
  url={https://openai.com/research/video-generation-models-as-world-simulators},
}

@misc{kling,
  title={Kling Video Generation},
  author={Kuaishou, Inc},
  year={2024},
  url={https://klingai.com/},
}

@misc{kingma2022vae,
      title={Auto-Encoding Variational Bayes}, 
      author={Diederik P Kingma and Max Welling},
      year={2022},
      eprint={1312.6114},
      archivePrefix={arXiv},
      primaryClass={stat.ML},
      url={https://arxiv.org/abs/1312.6114}, 
}

@misc{ho2020denoisingdiffusionprobabilisticmodels,
      title={Denoising Diffusion Probabilistic Models}, 
      author={Jonathan Ho and Ajay Jain and Pieter Abbeel},
      year={2020},
      eprint={2006.11239},
      archivePrefix={arXiv},
      primaryClass={cs.LG},
      url={https://arxiv.org/abs/2006.11239}, 
}

@misc{3dunet,
      title={U-Net: Convolutional Networks for Biomedical Image Segmentation}, 
      author={Olaf Ronneberger and Philipp Fischer and Thomas Brox},
      year={2015},
      eprint={1505.04597},
      archivePrefix={arXiv},
      primaryClass={cs.CV},
      url={https://arxiv.org/abs/1505.04597}, 
}

@misc{blattmann2023stablevideodiffusionscaling,
      title={Stable Video Diffusion: Scaling Latent Video Diffusion Models to Large Datasets}, 
      author={Andreas Blattmann and Tim Dockhorn and Sumith Kulal and Daniel Mendelevitch and Maciej Kilian and Dominik Lorenz and Yam Levi and Zion English and Vikram Voleti and Adam Letts and Varun Jampani and Robin Rombach},
      year={2023},
      eprint={2311.15127},
      archivePrefix={arXiv},
      primaryClass={cs.CV},
      url={https://arxiv.org/abs/2311.15127}, 
}

@misc{wang2023modelscope,
      title={ModelScope Text-to-Video Technical Report}, 
      author={Jiuniu Wang and Hangjie Yuan and Dayou Chen and Yingya Zhang and Xiang Wang and Shiwei Zhang},
      year={2023},
      eprint={2308.06571},
      archivePrefix={arXiv},
      primaryClass={cs.CV},
      url={https://arxiv.org/abs/2308.06571}, 
}

@article{Peebles2022DiT,
  title={Scalable Diffusion Models with Transformers},
  author={William Peebles and Saining Xie},
  year={2022},
  journal={arXiv preprint arXiv:2212.09748},
}

@article{qwen2.5,
    title   = {Qwen2.5 Technical Report}, 
    author  = {An Yang and Baosong Yang and Beichen Zhang and Binyuan Hui and Bo Zheng and Bowen Yu and Chengyuan Li and Dayiheng Liu and Fei Huang and Haoran Wei and Huan Lin and Jian Yang and Jianhong Tu and Jianwei Zhang and Jianxin Yang and Jiaxi Yang and Jingren Zhou and Junyang Lin and Kai Dang and Keming Lu and Keqin Bao and Kexin Yang and Le Yu and Mei Li and Mingfeng Xue and Pei Zhang and Qin Zhu and Rui Men and Runji Lin and Tianhao Li and Tingyu Xia and Xingzhang Ren and Xuancheng Ren and Yang Fan and Yang Su and Yichang Zhang and Yu Wan and Yuqiong Liu and Zeyu Cui and Zhenru Zhang and Zihan Qiu},
    journal = {arXiv preprint arXiv:2412.15115},
    year    = {2024}
}

@article{lipman2022flow,
  title={Flow matching for generative modeling},
  author={Lipman, Yaron and Chen, Ricky TQ and Ben-Hamu, Heli and Nickel, Maximilian and Le, Matt},
  journal={arXiv preprint arXiv:2210.02747},
  year={2022}
}

@inproceedings{evans2025stableaudioopen,
  title={Stable audio open},
  author={Evans, Zach and Parker, Julian D and Carr, CJ and Zukowski, Zack and Taylor, Josiah and Pons, Jordi},
  booktitle={ICASSP 2025-2025 IEEE International Conference on Acoustics, Speech and Signal Processing (ICASSP)},
  pages={1--5},
  year={2025},
  organization={IEEE}
}

@inproceedings{cheng2025lova,
  title={Lova: Long-form video-to-audio generation},
  author={Cheng, Xin and Wang, Xihua and Wu, Yihan and Wang, Yuyue and Song, Ruihua},
  booktitle={ICASSP 2025-2025 IEEE International Conference on Acoustics, Speech and Signal Processing (ICASSP)},
  pages={1--5},
  year={2025},
  organization={IEEE}
}

@InProceedings{wang2025jointdit,
    author    = {Wang, Xihua and Song, Ruihua and Li, Chongxuan and Cheng, Xin and Li, Boyuan and Wu, Yihan and Wang, Yuyue and Xu, Hongteng and Wang, Yunfeng},
    title     = {Animate and Sound an Image},
    booktitle = {Proceedings of the IEEE/CVF Conference on Computer Vision and Pattern Recognition (CVPR)},
    month     = {June},
    year      = {2025},
    pages     = {23369-23378}
}

@article{zhao2025uniform,
  title={UniForm: A Unified Multi-Task Diffusion Transformer for Audio-Video Generation},
  author={Zhao, Lei and Feng, Linfeng and Ge, Dongxu and Chen, Rujin and Yi, Fangqiu and Zhang, Chi and Zhang, Xiao-Lei and Li, Xuelong},
  journal={arXiv preprint arXiv:2502.03897},
  year={2025}
}

@inproceedings{wang2024tiva,
  title={Tiva: Time-aligned video-to-audio generation},
  author={Wang, Xihua and Wang, Yuyue and Wu, Yihan and Song, Ruihua and Tan, Xu and Chen, Zehua and Xu, Hongteng and Sui, Guodong},
  booktitle={Proceedings of the 32nd ACM International Conference on Multimedia},
  pages={573--582},
  year={2024}
}

@inproceedings{cheng2025mmaudio,
  title={MMAudio: Taming Multimodal Joint Training for High-Quality Video-to-Audio Synthesis},
  author={Cheng, Ho Kei and Ishii, Masato and Hayakawa, Akio and Shibuya, Takashi and Schwing, Alexander and Mitsufuji, Yuki},
  booktitle={Proceedings of the Computer Vision and Pattern Recognition Conference},
  pages={28901--28911},
  year={2025}
}

@inproceedings{xing2024seeing,
  title={Seeing and hearing: Open-domain visual-audio generation with diffusion latent aligners},
  author={Xing, Yazhou and He, Yingqing and Tian, Zeyue and Wang, Xintao and Chen, Qifeng},
  booktitle={Proceedings of the IEEE/CVF Conference on Computer Vision and Pattern Recognition},
  pages={7151--7161},
  year={2024}
}

@article{liu2025javisdit,
  title={Javisdit: Joint audio-video diffusion transformer with hierarchical spatio-temporal prior synchronization},
  author={Liu, Kai and Li, Wei and Chen, Lai and Wu, Shengqiong and Zheng, Yanhao and Ji, Jiayi and Zhou, Fan and Jiang, Rongxin and Luo, Jiebo and Fei, Hao and others},
  journal={arXiv preprint arXiv:2503.23377},
  year={2025}
}

@article{ho2022cfg,
  title={Classifier-free diffusion guidance},
  author={Ho, Jonathan and Salimans, Tim},
  journal={arXiv preprint arXiv:2207.12598},
  year={2022}
}

@article{liu2023audioldm,
  title={Audioldm: Text-to-audio generation with latent diffusion models},
  author={Liu, Haohe and Chen, Zehua and Yuan, Yi and Mei, Xinhao and Liu, Xubo and Mandic, Danilo and Wang, Wenwu and Plumbley, Mark D},
  journal={arXiv preprint arXiv:2301.12503},
  year={2023}
}

@article{liu2024audioldm,
  title={Audioldm 2: Learning holistic audio generation with self-supervised pretraining},
  author={Liu, Haohe and Yuan, Yi and Liu, Xubo and Mei, Xinhao and Kong, Qiuqiang and Tian, Qiao and Wang, Yuping and Wang, Wenwu and Wang, Yuxuan and Plumbley, Mark D},
  journal={IEEE/ACM Transactions on Audio, Speech, and Language Processing},
  volume={32},
  pages={2871--2883},
  year={2024},
  publisher={IEEE}
}

@article{liu2024syncflow,
  title={SyncFlow: Toward Temporally Aligned Joint Audio-Video Generation from Text},
  author={Liu, Haohe and Lan, Gael Le and Mei, Xinhao and Ni, Zhaoheng and Kumar, Anurag and Nagaraja, Varun and Wang, Wenwu and Plumbley, Mark D and Shi, Yangyang and Chandra, Vikas},
  journal={arXiv preprint arXiv:2412.15220},
  year={2024}
}

@article{wang2024avdit,
  title={Av-dit: Efficient audio-visual diffusion transformer for joint audio and video generation},
  author={Wang, Kai and Deng, Shijian and Shi, Jing and Hatzinakos, Dimitrios and Tian, Yapeng},
  journal={arXiv preprint arXiv:2406.07686},
  year={2024}
}

@article{tang2023codi,
  title={Any-to-any generation via composable diffusion},
  author={Tang, Zineng and Yang, Ziyi and Zhu, Chenguang and Zeng, Michael and Bansal, Mohit},
  journal={Advances in Neural Information Processing Systems},
  volume={36},
  pages={16083--16099},
  year={2023},
}

@inproceedings{jeong2023tpos,
  title={The power of sound (tpos): Audio reactive video generation with stable diffusion},
  author={Jeong, Yujin and Ryoo, Wonjeong and Lee, Seunghyun and Seo, Dabin and Byeon, Wonmin and Kim, Sangpil and Kim, Jinkyu},
  booktitle={Proceedings of the IEEE/CVF International Conference on Computer Vision},
  pages={7822--7832},
  year={2023}
}

@article{ishii2024ssvg,
  title={A simple but strong baseline for sounding video generation: Effective adaptation of audio and video diffusion models for joint generation},
  author={Ishii, Masato and Hayakawa, Akio and Shibuya, Takashi and Mitsufuji, Yuki},
  journal={arXiv preprint arXiv:2409.17550},
  year={2024}
}

@article{wan2025wan,
  title={Wan: Open and advanced large-scale video generative models},
  author={Wan, Team and Wang, Ang and Ai, Baole and Wen, Bin and Mao, Chaojie and Xie, Chen-Wei and Chen, Di and Yu, Feiwu and Zhao, Haiming and Yang, Jianxiao and others},
  journal={arXiv preprint arXiv:2503.20314},
  year={2025}
}

@misc{huang2023makeanaudio,
      title={Make-An-Audio 2: Temporal-Enhanced Text-to-Audio Generation}, 
      author={Jiawei Huang and Yi Ren and Rongjie Huang and Dongchao Yang and Zhenhui Ye and Chen Zhang and Jinglin Liu and Xiang Yin and Zejun Ma and Zhou Zhao},
      year={2023},
      eprint={2305.18474},
      archivePrefix={arXiv},
      primaryClass={cs.SD}
}

@article{karras2022edm,
  title={Elucidating the design space of diffusion-based generative models},
  author={Karras, Tero and Aittala, Miika and Aila, Timo and Laine, Samuli},
  journal={Advances in neural information processing systems},
  volume={35},
  pages={26565--26577},
  year={2022}
}

@article{wang2025klingaudio,
  title={Kling-Foley: Multimodal Diffusion Transformer for High-Quality Video-to-Audio Generation},
  author={Wang, Jun and Zeng, Xijuan and Qiang, Chunyu and Chen, Ruilong and Wang, Shiyao and Wang, Le and Zhou, Wangjing and Cai, Pengfei and Zhao, Jiahui and Li, Nan and others},
  journal={arXiv preprint arXiv:2506.19774},
  year={2025}
}

@inproceedings{ruan2022mmdiffusion,
author = {Ruan, Ludan and Ma, Yiyang and Yang, Huan and He, Huiguo and Liu, Bei and Fu, Jianlong and Yuan, Nicholas Jing and Jin, Qin and Guo, Baining},
title = {MM-Diffusion: Learning Multi-Modal Diffusion Models for Joint Audio and Video Generation},
year	= {2023},
booktitle	= {CVPR},
}

@inproceedings{sun2024mmldm,
  title={Mm-ldm: Multi-modal latent diffusion model for sounding video generation},
  author={Sun, Mingzhen and Wang, Weining and Qiao, Yanyuan and Sun, Jiahui and Qin, Zihan and Guo, Longteng and Zhu, Xinxin and Liu, Jing},
  booktitle={Proceedings of the 32nd ACM International Conference on Multimedia},
  pages={10853--10861},
  year={2024}
}

@inproceedings{lee2022landscape,
  title={Sound-Guided Semantic Video Generation},
  author={Lee, Seung Hyun and Oh, Gyeongrok and Byeon, Wonmin and Kim, Chanyoung and Ryoo, Won Jeong and Yoon, Sang Ho and Cho, Hyunjun and Bae, Jihyun and Kim, Jinkyu and Kim, Sangpil},
  booktitle={Computer Vision--ECCV 2022: 17th European Conference, Tel Aviv, Israel, October 23--27, 2022, Proceedings, Part XVII},
  pages={34--50},
  year={2022},
  organization={Springer}
}

@article{weng2025mtv,
  title={Audio-Sync Video Generation with Multi-Stream Temporal Control},
  author={Weng, Shuchen and Zheng, Haojie and Chang, Zheng and Li, Si and Shi, Boxin and Wang, Xinlong},
  journal={arXiv preprint arXiv:2506.08003},
  year={2025}
}

@misc{zhang2023controlnet,
  title={Adding Conditional Control to Text-to-Image Diffusion Models}, 
  author={Lvmin Zhang and Anyi Rao and Maneesh Agrawala},
  booktitle={IEEE International Conference on Computer Vision (ICCV)},
  year={2023},
}

@article{Qwen2.5-VL,
  title={Qwen2.5-VL Technical Report},
  author={Bai, Shuai and Chen, Keqin and Liu, Xuejing and Wang, Jialin and Ge, Wenbin and Song, Sibo and Dang, Kai and Wang, Peng and Wang, Shijie and Tang, Jun and Zhong, Humen and Zhu, Yuanzhi and Yang, Mingkun and Li, Zhaohai and Wan, Jianqiang and Wang, Pengfei and Ding, Wei and Fu, Zheren and Xu, Yiheng and Ye, Jiabo and Zhang, Xi and Xie, Tianbao and Cheng, Zesen and Zhang, Hang and Yang, Zhibo and Xu, Haiyang and Lin, Junyang},
  journal={arXiv preprint arXiv:2502.13923},
  year={2025}
}

@article{2020t5,
  author  = {Colin Raffel and Noam Shazeer and Adam Roberts and Katherine Lee and Sharan Narang and Michael Matena and Yanqi Zhou and Wei Li and Peter J. Liu},
  title   = {Exploring the Limits of Transfer Learning with a Unified Text-to-Text Transformer},
  journal = {Journal of Machine Learning Research},
  year    = {2020},
  volume  = {21},
  number  = {140},
  pages   = {1-67},
  url     = {http://jmlr.org/papers/v21/20-074.html}
}

@inproceedings{linz2024avsync,
    title={Audio-Synchronized Visual Animation},
    author={Lin Zhang and Shentong Mo and Yijing Zhang and Pedro Morgado},
    booktitle={Proceedings of the European Conference on Computer Vision (ECCV)},
    year={2024}
}

@InProceedings{Chen20vggsound,
  author       = "Honglie Chen and Weidi Xie and Andrea Vedaldi and Andrew Zisserman",
  title        = "VGGSound: A Large-scale Audio-Visual Dataset",
  booktitle    = "International Conference on Acoustics, Speech, and Signal Processing (ICASSP)",
  year         = "2020",
}

@InProceedings{Chen21vggss,
title="Localizing Visual Sounds the Hard Way",
author="Honglie Chen and Weidi Xie and Triantafyllos Afouras and Arsha Nagrani and Andrea Vedaldi and Andrew Zisserman",
booktitle="CVPR",
year="2021"}

@misc{wu2025qwenimagetechnicalreport,
      title={Qwen-Image Technical Report}, 
      author={Chenfei Wu and Jiahao Li and Jingren Zhou and Junyang Lin and Kaiyuan Gao and Kun Yan and Sheng-ming Yin and Shuai Bai and Xiao Xu and Yilei Chen and Yuxiang Chen and Zecheng Tang and Zekai Zhang and Zhengyi Wang and An Yang and Bowen Yu and Chen Cheng and Dayiheng Liu and Deqing Li and Hang Zhang and Hao Meng and Hu Wei and Jingyuan Ni and Kai Chen and Kuan Cao and Liang Peng and Lin Qu and Minggang Wu and Peng Wang and Shuting Yu and Tingkun Wen and Wensen Feng and Xiaoxiao Xu and Yi Wang and Yichang Zhang and Yongqiang Zhu and Yujia Wu and Yuxuan Cai and Zenan Liu},
      year={2025},
      eprint={2508.02324},
      archivePrefix={arXiv},
      primaryClass={cs.CV},
      url={https://arxiv.org/abs/2508.02324}, 
}

@article{unterthiner2018fvd,
  title={Towards accurate generative models of video: A new metric \& challenges},
  author={Unterthiner, Thomas and Van Steenkiste, Sjoerd and Kurach, Karol and Marinier, Raphael and Michalski, Marcin and Gelly, Sylvain},
  journal={arXiv preprint arXiv:1812.01717},
  year={2018}
}

@article{kilgour2018fad,
  title={Fr$\backslash$'echet audio distance: A metric for evaluating music enhancement algorithms},
  author={Kilgour, Kevin and Zuluaga, Mauricio and Roblek, Dominik and Sharifi, Matthew},
  journal={arXiv preprint arXiv:1812.08466},
  year={2018}
}

@inproceedings{radford2021clip,
  title={Learning transferable visual models from natural language supervision},
  author={Radford, Alec and Kim, Jong Wook and Hallacy, Chris and Ramesh, Aditya and Goh, Gabriel and Agarwal, Sandhini and Sastry, Girish and Askell, Amanda and Mishkin, Pamela and Clark, Jack and others},
  booktitle={International conference on machine learning},
  pages={8748--8763},
  year={2021},
  organization={PmLR}
}

@inproceedings{elizalde2023clap,
  title={Clap learning audio concepts from natural language supervision},
  author={Elizalde, Benjamin and Deshmukh, Soham and Al Ismail, Mahmoud and Wang, Huaming},
  booktitle={ICASSP 2023-2023 IEEE International Conference on Acoustics, Speech and Signal Processing (ICASSP)},
  pages={1--5},
  year={2023},
  organization={IEEE}
}

@inproceedings{girdhar2023imagebind,
  title={ImageBind: One Embedding Space To Bind Them All},
  author={Girdhar, Rohit and El-Nouby, Alaaeldin and Liu, Zhuang
and Singh, Mannat and Alwala, Kalyan Vasudev and Joulin, Armand and Misra, Ishan},
  booktitle={CVPR},
  year={2023}
}

@inproceedings{yariv2024avalign,
  title={Diverse and aligned audio-to-video generation via text-to-video model adaptation},
  author={Yariv, Guy and Gat, Itai and Benaim, Sagie and Wolf, Lior and Schwartz, Idan and Adi, Yossi},
  booktitle={Proceedings of the AAAI Conference on Artificial Intelligence},
  volume={38},
  pages={6639--6647},
  year={2024}
}

@article{Qwen2-Audio,
  title={Qwen2-Audio Technical Report},
  author={Chu, Yunfei and Xu, Jin and Yang, Qian and Wei, Haojie and Wei, Xipin and Guo,  Zhifang and Leng, Yichong and Lv, Yuanjun and He, Jinzheng and Lin, Junyang and Zhou, Chang and Zhou, Jingren},
  journal={arXiv preprint arXiv:2407.10759},
  year={2024}
}

@article{salimans2022vpredict,
  title={Progressive distillation for fast sampling of diffusion models},
  author={Salimans, Tim and Ho, Jonathan},
  journal={arXiv preprint arXiv:2202.00512},
  year={2022}
}

@article{hacohen2026ltx,
  title={LTX-2: Efficient Joint Audio-Visual Foundation Model},
  author={HaCohen, Yoav and Brazowski, Benny and Chiprut, Nisan and Bitterman, Yaki and Kvochko, Andrew and Berkowitz, Avishai and Shalem, Daniel and Lifschitz, Daphna and Moshe, Dudu and Porat, Eitan and others},
  journal={arXiv preprint arXiv:2601.03233},
  year={2026}
}

@misc{low2025ovi,
      title={Ovi: Twin Backbone Cross-Modal Fusion for Audio-Video Generation}, 
      author={Chetwin Low and Weimin Wang and Calder Katyal},
      year={2025},
      eprint={2510.01284},
      archivePrefix={arXiv},
      primaryClass={cs.MM},
      url={https://arxiv.org/abs/2510.01284}, 
}

@misc{esser2024mmdit,
      title={Scaling Rectified Flow Transformers for High-Resolution Image Synthesis}, 
      author={Patrick Esser and Sumith Kulal and Andreas Blattmann and Rahim Entezari and Jonas Müller and Harry Saini and Yam Levi and Dominik Lorenz and Axel Sauer and Frederic Boesel and Dustin Podell and Tim Dockhorn and Zion English and Kyle Lacey and Alex Goodwin and Yannik Marek and Robin Rombach},
      year={2024},
      eprint={2403.03206},
      archivePrefix={arXiv},
      primaryClass={cs.CV},
      url={https://arxiv.org/abs/2403.03206}, 
}

@misc{carreira2018i3d,
      title={Quo Vadis, Action Recognition? A New Model and the Kinetics Dataset}, 
      author={Joao Carreira and Andrew Zisserman},
      year={2018},
      eprint={1705.07750},
      archivePrefix={arXiv},
      primaryClass={cs.CV},
      url={https://arxiv.org/abs/1705.07750}, 
}

@InProceedings{Guan_2025_ICCV,
    author    = {Guan, Kaisi and Lai, Zhengfeng and Sun, Yuchong and Zhang, Peng and Liu, Wei and Liu, Kieran and Cao, Meng and Song, Ruihua},
    title     = {ETVA: Evaluation of Text-to-Video Alignment via Fine-grained Question Generation and Answering},
    booktitle = {Proceedings of the IEEE/CVF International Conference on Computer Vision (ICCV)},
    month     = {October},
    year      = {2025},
    pages     = {21299-21309}
}

@misc{vssflow,
title = {VSSFlow: Unifying Video-conditioned Sound and Speech Generation via Joint Learning},
author = {Xin Cheng and Yuyue Wang and Xihua Wang and Yihan Wu and Kaisi Guan and Yijing Chen and Peng Zhang and Kieran Liu and Meng Cao and Ruihua Song},
year = {2026},
URL = {https://www.arxiv.org/abs/2509.24773}
}

@misc{chen2025detectingmitigatinginsertionhallucination,
      title={Detecting and Mitigating Insertion Hallucination in Video-to-Audio Generation}, 
      author={Liyang Chen and Hongkai Chen and Yujun Cai and Sifan Li and Qingwen Ye and Yiwei Wang},
      year={2025},
      eprint={2510.08078},
      archivePrefix={arXiv},
      primaryClass={cs.SD},
      url={https://arxiv.org/abs/2510.08078}, 
}

@inproceedings{
    chen2025sanavideo,
    title={SANA-Video: Efficient Video Generation with Block Linear Diffusion Transformer}, 
    author={Junsong Chen and Yuyang Zhao and Jincheng Yu and Ruihang Chu and Junyu Chen and Shuai Yang and Xianbang Wang and Yicheng Pan and Daquan Zhou and Huan Ling and Haozhe Liu and Hongwei Yi and Hao Zhang and Muyang Li and Yukang Chen and Han Cai and Sanja Fidler and Ping Luo and Song Han and Enze Xie},
    booktitle={The Fourteenth International Conference on Learning Representations},
    year={2026},
    url={https://openreview.net/forum?id=mzAchylAtf}
}

@misc{liu2025thinksoundchainofthoughtreasoningmultimodal,
      title={ThinkSound: Chain-of-Thought Reasoning in Multimodal Large Language Models for Audio Generation and Editing}, 
      author={Huadai Liu and Jialei Wang and Kaicheng Luo and Wen Wang and Qian Chen and Zhou Zhao and Wei Xue},
      year={2025},
      eprint={2506.21448},
      archivePrefix={arXiv},
      primaryClass={eess.AS},
      url={https://arxiv.org/abs/2506.21448}, 
}

@misc{liu2025prismaudiodecomposedchainofthoughtsmultidimensional,
          title={PrismAudio: Decomposed Chain-of-Thoughts and Multi-dimensional Rewards for Video-to-Audio Generation}, 
          author={Huadai Liu and Kaicheng Luo and Wen Wang and Qian Chen and Peiwen Sun and Rongjie Huang and Xiangang Li and Jieping Ye and Wei Xue},
          year={2025},
          eprint={2511.18833},
          archivePrefix={arXiv},
          primaryClass={cs.SD},
          url={https://arxiv.org/abs/2511.18833}, 
    }

@article{chung2023unimax,
  title={Unimax: Fairer and more effective language sampling for large-scale multilingual pretraining},
  author={Chung, Hyung Won and Constant, Noah and Garcia, Xavier and Roberts, Adam and Tay, Yi and Narang, Sharan and Firat, Orhan},
  journal={arXiv preprint arXiv:2304.09151},
  year={2023}
}

\newpage
\appendix

\section{Diffusion and Flow Matching Generation Models}
\label{app:back}
Generative models are designed to learn a complex data distribution $p(\mathbf{x})$ from a simple prior, typically a standard Gaussian distribution $\mathcal{N}(\mathbf{0}, \mathbf{I})$. Many state-of-the-art approaches are based on learning to reverse a predefined process that maps data to noise. A prominent family of such models is diffusion models. In their foundational formulation (DDPM)~\citep{ho2020denoisingdiffusionprobabilisticmodels}, they utilize a fixed forward process that progressively adds Gaussian noise to a data sample $\mathbf{x}_0$ over discrete timesteps. The resulting noisy sample at any time $t$, denoted $\mathbf{x}_t$, can be expressed as $\mathbf{x}_t = \sqrt{\bar{\alpha}_t}\mathbf{x}_0 + \sqrt{1 - \bar{\alpha}_t}\boldsymbol{\epsilon}$, where $\bar{\alpha}_t$ is a predefined noise schedule and $\boldsymbol{\epsilon} \sim \mathcal{N}(\mathbf{0}, \mathbf{I})$. A neural network, $\boldsymbol{\epsilon}_\theta(\mathbf{x}_t, t)$, is then trained to predict the noise component $\boldsymbol{\epsilon}$ from the corrupted sample:
\begin{equation*}
L_{\text{DDPM}}(\theta) = \mathbb{E}_{t, \mathbf{x}_0, \boldsymbol{\epsilon}} \left[ || \boldsymbol{\epsilon} - \boldsymbol{\epsilon}_\theta(\mathbf{x}_t, t) ||^2 \right]
\end{equation*}
More recent frameworks, such as EDM~\citep{karras2022edm}, generalize this process by formulating it as solving a continuous-time stochastic differential equation (SDE). EDM provides a principled design methodology, emphasizing crucial choices like network preconditioning. The denoiser network, $D_\theta(\mathbf{x}_t, \sigma_t)$, is scaled to have consistent input and output magnitudes across all noise levels $\sigma_t$. This network is often trained to predict the clean data $\mathbf{x}_0$ directly, using a weighted loss function that prioritizes different noise levels:
\begin{equation*}
L_{\text{EDM}}(\theta) = \mathbb{E}_{t, \mathbf{x}_0, \boldsymbol{\epsilon}} \left[ \lambda(\sigma_t) || D_\theta(\mathbf{x}_t, \sigma_t) - \mathbf{x}_0 ||^2 \right]
\end{equation*}
For the generation part, both approaches start with a sample from the prior, $\mathbf{x}_T \sim \mathcal{N}(\mathbf{0}, \mathbf{I})$, and iteratively apply the learned denoising function to recover a clean sample $\mathbf{x}_0$.
As an alternative to the noise-prediction framework, Flow Matching (FM)~\citep{lipman2022flow} models learn to generate data in a single continuous-time transformation. These models learn a vector field $\mathbf{v}_t$ that transports samples from a prior distribution $p_0$ (noise) to the target data distribution $p_1$ (data) by following an ordinary differential equation (ODE): $\frac{d\mathbf{x}_t}{dt} = \mathbf{v}_t(\mathbf{x}_t)$. To make training tractable, FM trains a network $v_\theta$ to approximate a simple, predefined vector field. For a linear path between a noise sample $\mathbf{x}_0 \sim p_0$ and a data sample $\mathbf{x}_1 \sim p_1$, the target vector field is simply their difference, $\mathbf{x}_1 - \mathbf{x}_0$. The corresponding FM loss is:
\begin{equation*}
L_{\text{FM}}(\theta) = \mathbb{E}_{t, \mathbf{x}_0, \mathbf{x}_1} \left[ || v_\theta(t, t\mathbf{x}_0 + (1-t)\mathbf{x}_1) - (\mathbf{x}_1 - \mathbf{x}_0) ||^2 \right]
\end{equation*}
To generate a sample, one simply solves the learned ODE $\frac{d\mathbf{x}_t}{dt} = v_\theta(t, \mathbf{x}_t)$ from $t=0$ to $t=1$, starting with an initial sample $\mathbf{x}_0 \sim p_0$.

\paragraph{Classifier-Free Guidance} Conditional generation in these models is commonly achieved using Classifier-Free Guidance (CFG)~\citep{ho2022cfg}. This technique steers the generation process towards a desired condition $c$ (e.g., a text prompt) without needing an external classifier. The model, here denoted with the noise predictor $\boldsymbol{\epsilon}_\theta(\mathbf{x}_t, t, c)$, is jointly trained on conditional inputs $c$ and a null token $\emptyset$. During sampling, the guided prediction $\hat{\boldsymbol{\epsilon}}_\theta$ is an extrapolation from the unconditional prediction towards the conditional one:
\begin{equation*}
\hat{\boldsymbol{\epsilon}}_\theta = \boldsymbol{\epsilon}_\theta(\mathbf{x}_t, t, \emptyset) + w (\boldsymbol{\epsilon}_\theta(\mathbf{x}_t, t, c) - \boldsymbol{\epsilon}_\theta(\mathbf{x}_t, t, \emptyset))
\end{equation*}
The guidance scale $w > 1$ is a hyperparameter that adjusts the strength of the condition. A larger $w$ typically improves fidelity to the condition at the cost of reduced sample diversity. This technique is applied analogously to other model predictions like $D_\theta$ or $v_\theta$.

\section{Experimental Details}
\label{app:exp_setup}
\subsection{Compute Resources}
\label{app:com}
All experiments were conducted on 4 NVIDIA H100 80GB GPUs. We use the DeepSpeed ZeRO-2 framework for distributed training, enabling efficient memory management across GPUs. Each node utilized 64 Intel(R) Xeon(R) Platinum 8481C CPUs @ 2.70GHz, with 2TB of RAM and 4TB of SSD storage. Training on the AVSync15 dataset takes approximately 8 hours under this configuration. All quantitative results reported in this paper are obtained under this setup.
\subsection{Baselines}
\label{app:baselines}
Here we detail the baseline models used in our work.
\begin{itemize}[leftmargin=0.5cm]
    \item \textbf{Wan}~\citep{wan2025wan} is a large-scale video generative model (available in 1.3B and 14B versions) renowned for producing high-resolution and temporally coherent videos, representing a leading open-source T2V model.
    \item \textbf{Stable-Audio-Open}~\citep{evans2025stableaudioopen} is a diffusion-based text-to-audio generation model trained on a large dataset to create diverse and realistic audio content.
    \item \textbf{MMAudio}~\citep{cheng2025mmaudio} is a video-to-audio synthesis model designed to generate synchronized sound for silent video clips.
    \item \textbf{Seeing-and-Hearing}~\citep{xing2024seeing} introduces "diffusion latent aligners" that leverage the ImageBind embedding space to create a shared latent space for visual and auditory data, enabling semantic alignment guidance.
    \item \textbf{ThinkSound}~\citep{liu2025thinksoundchainofthoughtreasoningmultimodal} is a Chain-of-Thought-guided video-to-audio generation and editing framework. It leverages multimodal large language models to reason about visual dynamics, acoustic environments, and temporal relationships, guiding a unified audio foundation model to generate semantically aligned and temporally synchronized sounds.
    \item \textbf{PrismAudio}~\citep{liu2025prismaudiodecomposedchainofthoughtsmultidimensional} is a video-to-audio generation framework that decomposes audio-visual reasoning into semantic, temporal, aesthetic, and spatial Chain-of-Thought modules. It further introduces multi-dimensional reward optimization with Fast-GRPO to improve audio quality, synchronization, and spatial consistency.
    \item \textbf{TPos}~\citep{jeong2023tpos} focuses on audio-reactive video generation, creating dynamic and visually engaging videos that respond to the rhythm and emotional tone of an input audio track.
    \item \textbf{TempoToken}~\citep{yariv2024avalign} proposes "TempoTokens," learnable embeddings that guide audio-to-video generation, ensuring both temporal alignment between audio and visual output.
    \item \textbf{JointDiT}~\citep{wang2025jointdit} is a dual-tower joint generative model for image-conditioned sound video generation. It employs a Full Attention fusion mechanism, though its performance can be limited by its T2V backbone (e.g., Stable Video Diffusion).
    \item \textbf{JavisDiT}~\citep{liu2025javisdit} is a Joint Audio-Video Diffusion Transformer (JAVG) built on the DiT architecture. It achieves high-quality, synchronized audio-video generation from open-ended prompts by introducing a Hierarchical Spatio-Temporal Synchronized Prior (HiST-Sypo) Estimator for fine-grained alignment. 
    \item \textbf{SSVG}~\citep{ishii2024ssvg} presents a simple yet strong baseline for sounding video generation. It integrates base audio and video diffusion models with novel mechanisms like timestep adjustment and Cross-Modal Conditioning as Positional Encoding (CMC-PE), which is an additive-fusion mechanism to enhance audio-video alignment.
    \item \textbf{MTV}~\citep{weng2025mtv} is a versatile framework for audio-sync video generation that explicitly separates audio into speech, effects, and music tracks. This enables disentangled control over lip motion, event timing, and visual mood, leading to fine-grained and semantically aligned video generation. It also introduces the DEMIX dataset.
    \item \textbf{CoDi}~\citep{tang2023codi} (Composable Diffusion) is a versatile any-to-any generation model that composes diffusion models trained on different modalities to handle various input and output modalities, including text, images, video, and audio.
\end{itemize}
\subsection{Inference of Our BridgeDiT Model}
\label{app:infer}
For both video and audio generation, we apply Classifier-Free Guidance independently, leveraging separate guidance scales for each modality to fine-tune their respective generation quality and adherence to the text prompts. The guided noise prediction for each modality is given by:
\begin{align}
\hat{\epsilon}_v(\mathbf{x}_v, T_V) &= \epsilon_v(\mathbf{x}_v, \emptyset) + w_v \cdot (\epsilon_v(\mathbf{x}_v, T_V) - \epsilon_v(\mathbf{x}_v, \emptyset)) \\
\hat{\epsilon}_a(\mathbf{x}_a, T_A) &= \epsilon_a(\mathbf{x}_a, \emptyset) + w_a \cdot (\epsilon_a(\mathbf{x}_a, T_A) - \epsilon_a(\mathbf{x}_a, \emptyset))
\end{align}
Here, $\mathbf{x}_v$ and $\mathbf{x}_a$ represent the noisy video and audio latents at a given timestep, respectively. $\epsilon_v(\mathbf{x}_v, T_V)$ and $\epsilon_a(\mathbf{x}_a, T_A)$ are the predictions from the BridgeDiT model conditioned on their respective text prompts, while $\epsilon_v(\mathbf{x}_v, \emptyset)$ and $\epsilon_a(\mathbf{x}_a, \emptyset)$ are predictions from unconditioned (null) prompts. $w_v$ and $w_a$ are the video and audio guidance scales, allowing for independent control over the trade-off between sample quality and text alignment for each modality.
\begin{figure}[!t]
    \centering
    \includegraphics[width=\linewidth]{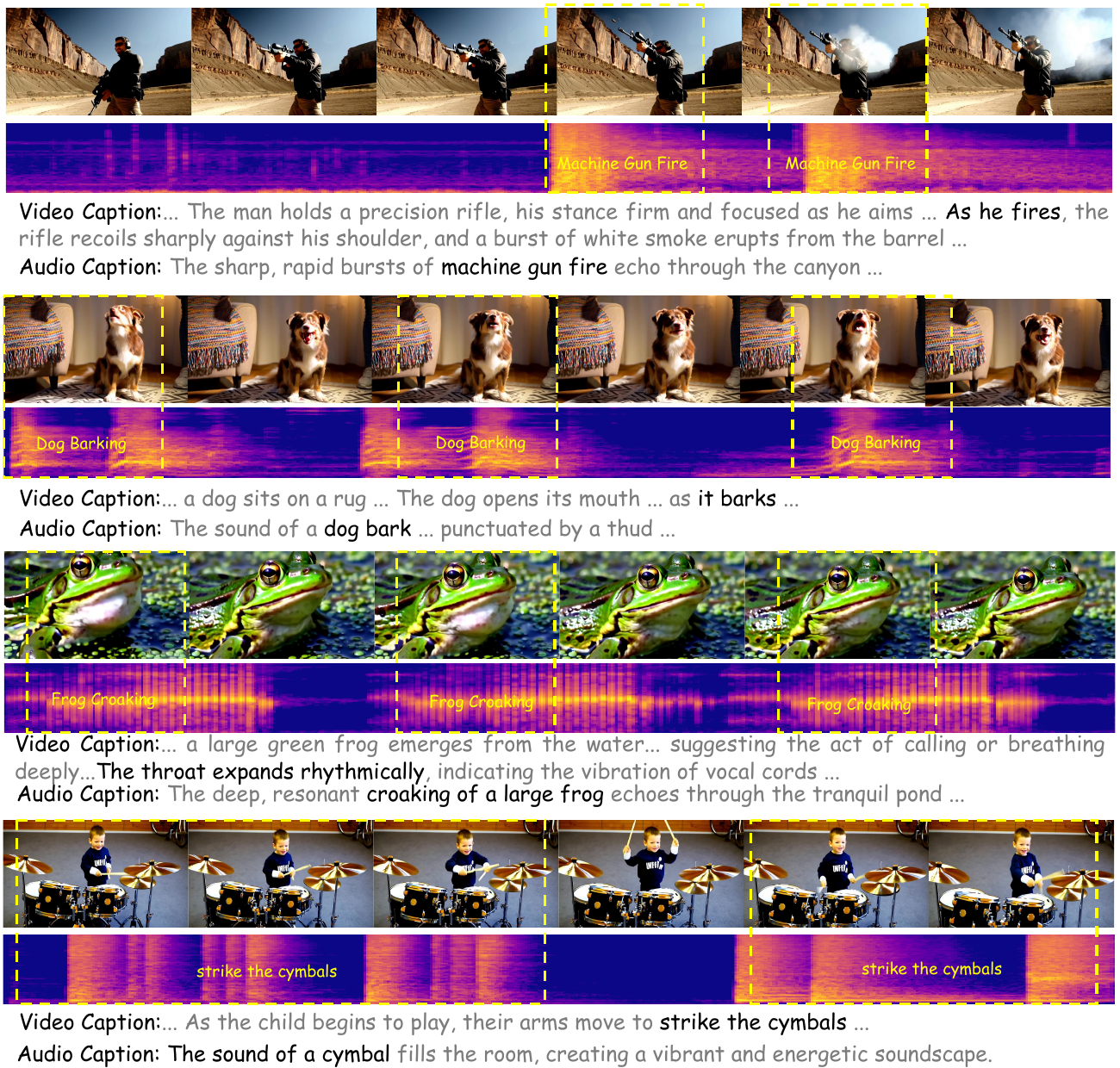}
    \caption{More Case Study}
    \label{app:case_study_fig}
\end{figure}
\subsection{More Case Study}
\label{app:case_study}
More Case Studies are in Fig~\ref{app:case_study_fig}.
\subsection{Hyperparameters}
Hyperparameters details are shown in Table~\ref{app:hyper}.
\begin{table}[ht!]
\centering
\caption{Key hyperparameters for our BridgeDiT model.}
\label{tab:hyperparameters}
\begin{tabular}{@{}ll@{}}
\toprule
\textbf{Parameter} & \textbf{Value} \\
\midrule
\addlinespace
\multicolumn{2}{l}{\textit{\textbf{Training Configuration}}} \\
\addlinespace
\hline
\addlinespace
Optimizer & AdamW \\
Learning Rate & 5e-5 \\
Weight Decay & 1e-3 \\
Adam Betas & (0.9, 0.95) \\
LR Scheduler & Cosine decay with linear warmup \\
LR Warmup Steps & 1,000 \\
Total Training Steps & 15,000 \\
Minimum Learning Rate & 1e-6 \\
Unconditional Probability (CFG) & 0.1 \\
Training Precision & bfloat16 \\
\addlinespace
\hline
\addlinespace
\multicolumn{2}{l}{\textit{\textbf{Architecture Configuration}}} \\
\addlinespace
\hline
\addlinespace
Number of BridgeDiT Blocks & 4 \\
BridgeDiT Block Channels (Q, K, V) & 1536 \\
BridgeDiT Block Heads & 12 \\
BridgeDiT Timestep Embedding Dim & 1536 \\
Video Tower Bridge Points (Layers) & [3, 7, 11, 15] \\
Audio Tower Bridge Points (Layers) & [2, 5, 8, 11] \\
Trainable Layers (Video) & Last 5 blocks \\
Trainable Layers (Audio) & Last 5 blocks \\
\addlinespace
\hline
\addlinespace
\multicolumn{2}{l}{\textit{\textbf{Sampling Configuration}}} \\
\addlinespace
\hline
\addlinespace
Video Resolution & 832 $\times$ 480 \\
Video Number of Frames & 81 \\
Video Frame Rate (fps) & 15 \\
Audio Sample Rate & 44100 Hz \\
Audio Duration & 5.4 seconds \\
Number of Inference Steps & 50 \\
Video Guidance Scale (CFG) & 6.0 \\
Audio Guidance Scale (CFG) & 6.0 \\
\addlinespace
\bottomrule
\end{tabular}
\label{app:hyper}
\end{table}
\subsection{Detail prompts for CRR Caption Framework}
\label{app:prompt}
\begin{figure}[!t]
    \centering
\includegraphics[width=\linewidth]{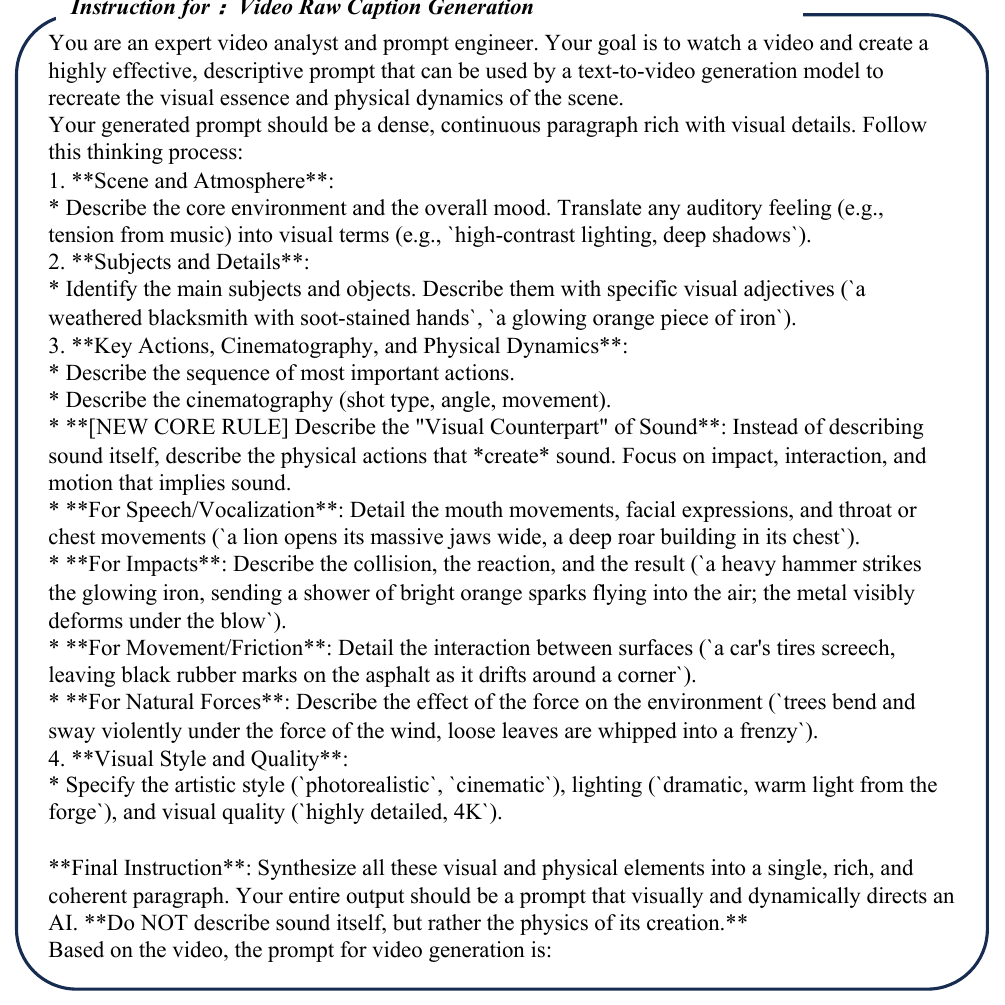}
\vspace*{-0.5cm}
\end{figure}
\begin{figure}[!t]
    \centering
\includegraphics[width=\linewidth]{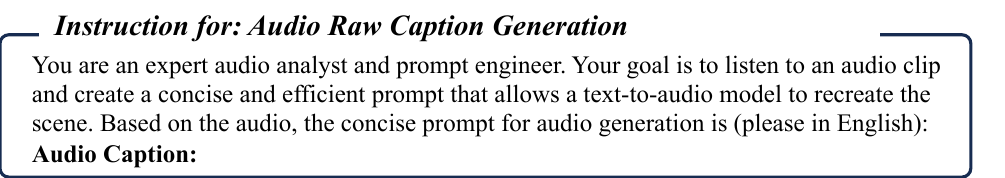}
\vspace*{-0.5cm}
\end{figure}
\clearpage
\begin{figure}[!t]
    \centering
\includegraphics[width=\linewidth]{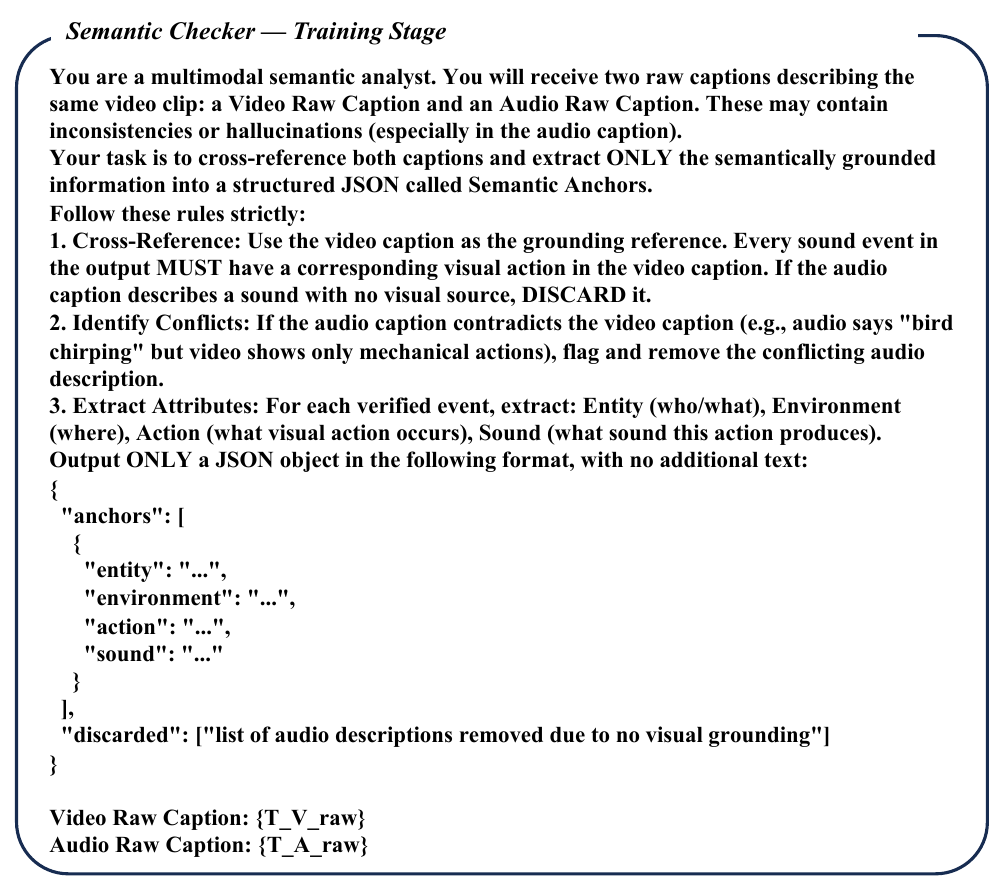}
\vspace*{-0.5cm}
\label{fig:step3}
\end{figure}
\begin{figure}[!t]
    \centering
\includegraphics[width=\linewidth]{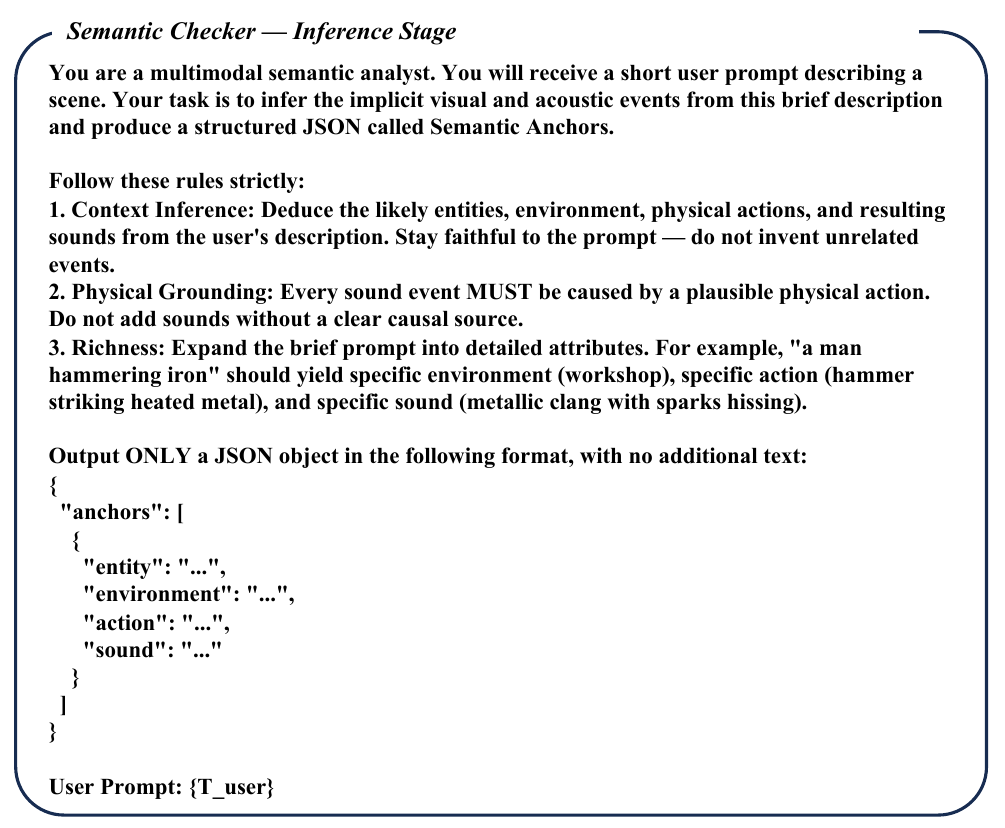}
\vspace*{-0.5cm}
\label{fig:step3}
\end{figure}
\begin{figure}[!t]
    \centering
\includegraphics[width=\linewidth]{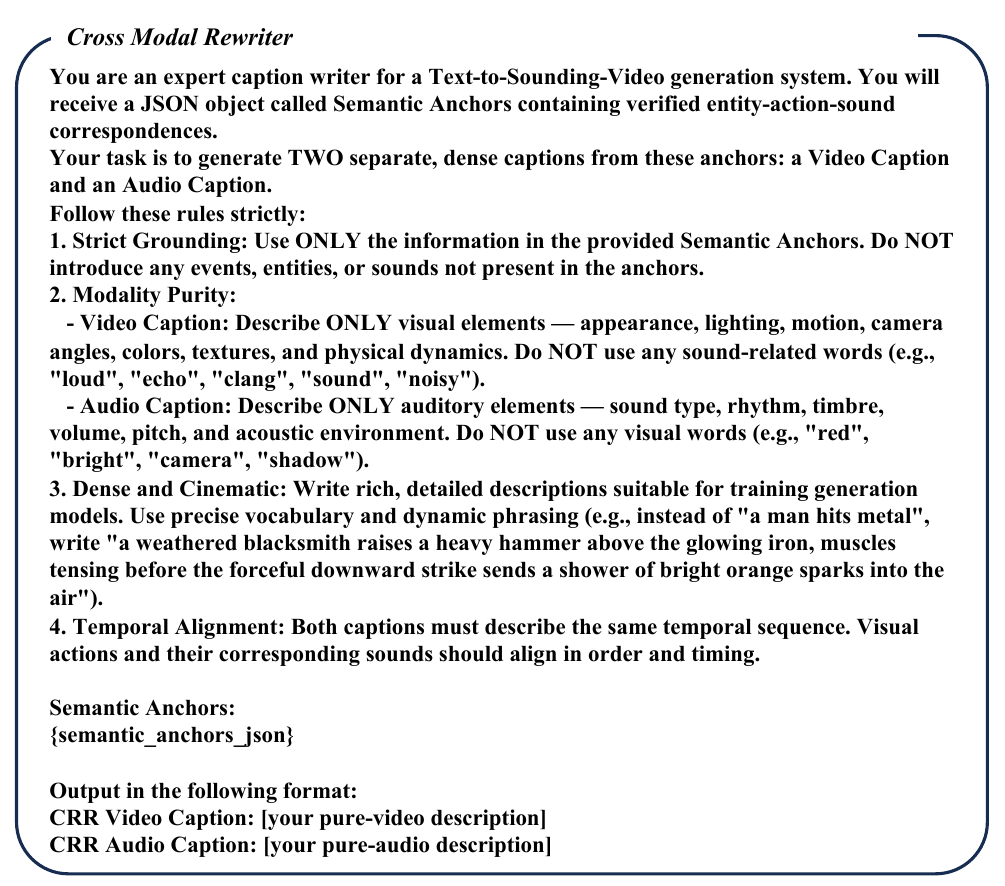}
\vspace*{-0.5cm}
\label{fig:step3}
\end{figure}
\clearpage
\subsection{Detailed Command for Human Annotation}
\label{app:human}
\begin{figure}[!t]
    \centering
\includegraphics[width=\linewidth]{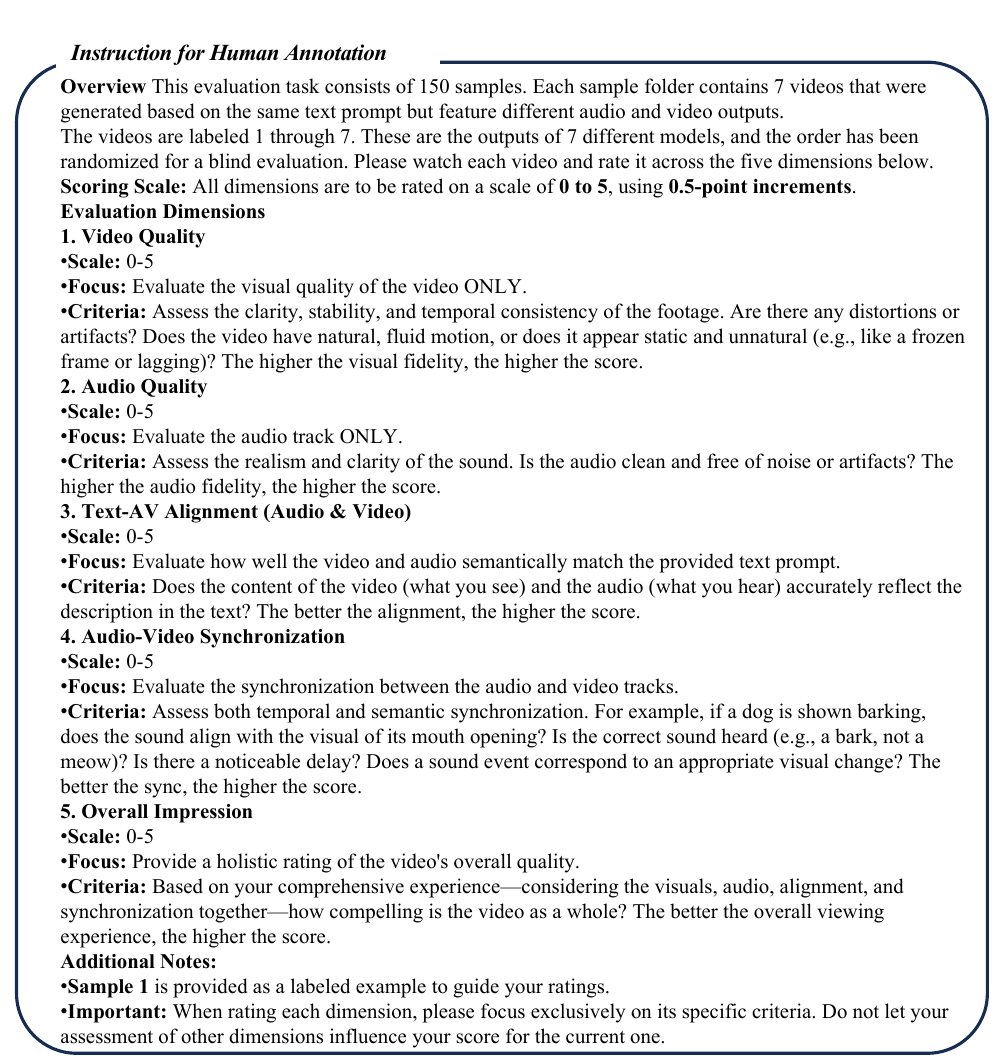}
\vspace*{-0.5cm}
\caption{Detailed Command for Human Annotation}
\label{fig:step4}
\end{figure}
\clearpage
\subsection{Examples Results for CRR Framework}
\begin{longtblr}[
caption={Examples of captions generated by our CRR Case1.},
label={app:example_caption}
]{
  width = \textwidth,
  colspec = {Q[l, m, font=\bfseries] X[l, m]},
  rowsep = 6pt,
}

\toprule
CRR Video Caption & In a dimly lit blacksmith workshop illuminated by the dramatic, warm light emanating from the forge, a weathered blacksmith with soot-stained hands stands at his anvil, his face etched with concentration. The atmosphere is thick with anticipation, the air filled with the scent of heated metal and the faint glow reflecting off the walls. The blacksmith, clad in a rugged leather apron, holds a glowing orange piece of iron with tongs, its intense heat casting flickering shadows across the room. He raises a heavy hammer above his head, muscles tensing as he prepares to strike. The hammer descends with force, colliding with the glowing iron and sending a shower of bright orange sparks flying into the air; the metal visibly deforms under the powerful blow. The blacksmith's arm recoils slightly from the impact, and he quickly adjusts his grip on the tongs, turning the iron to shape it further. Each strike is precise, the hammer connecting with the metal in a rhythmic dance of creation. The camera captures this intense moment from a close-up angle, focusing on the intricate details of the hammer's texture and the vibrant hues of the heated iron. The anvil, worn from years of use, bears the marks of countless creations, adding to the scene's authenticity. The visual counterpart of the hammer's resounding impacts is evident in the way the metal bends and reshapes, each deformation a testament to the blacksmith's skill. The entire scene is rendered in a photorealistic, cinematic style. \\
\midrule
CRR Audio Caption & The resounding clang of a heavy hammer striking metal echoes through the workshop, each powerful blow punctuated by the sizzle of heated iron and the occasional hiss of sparks. \\
\midrule
Video-raw Caption & A medium shot of a worker, perhaps a blacksmith or metal worker, standing in a workshop that is mostly dark except for some forge light. He is wearing protective gear like an apron and has some soot on his hands and face. He holds a glowing red piece of iron with pliers or tongs and stands near a big metal block. The room has heated metal smells and glows reflecting on the walls. The worker lifts a hammer and hits the metal block. Sparks fly into the air when the hammer descends and collides with the iron, which changes shape. His arm moves back a bit from the hit. The camera is close and shows the hammer and the iron. The metal object bends as he hits it. Realistic video style. \\
\midrule
Audio-raw Caption & There was a knocking sound, a bit like the sound of stones.\\ 
\midrule
Audio-Visual Caption & In a dark workshop illuminated by forge light reflecting on the walls, a worker wearing an apron with soot on his hands and face stands near a big metal block, holding a piece of glowing red iron with tongs. A close-up shot shows the worker lifting a hammer and hitting the metal object. Corresponding to the visual impact where sparks fly, and the iron bends, a knocking sound that is a bit like stones is heard. The scene smells of heated metal and is rendered in a realistic video style. \\ 
\bottomrule
\end{longtblr}

\clearpage

\section{Ablation Study on BridgeDiT Block Placement}
\label{app:placement}
\begin{table}[h!]
\centering
\vspace{-20pt}
\caption{Ablation study on the placement of BridgeDiT Blocks. Performance is highest when interaction is focused on the early-to-mid layers of the architecture.}
\label{tab:placement_ablation}
\begin{tabular}{lcccc}
\toprule
\textbf{Placement Strategy} & \textbf{Video Layers} & \textbf{Audio Layers} & \textbf{VA-IB ↑} & \textbf{AV-Align ↑} \\
\midrule
Early Layers & [0, 1, 2, 3] & [0, 1, 2, 3] & 28.30 & 0.2223 \\
Middle Layers & [13, 14, 15, 16] & [10, 11, 12, 13] & 31.89 & 0.2481 \\
Late Layers & [27, 28, 29, 30] & [21, 22, 23, 24] & 19.32 & 0.1831 \\
Uniform & [6, 12, 18, 24] & [2, 8, 13, 18] & 33.65 & 0.2502 \\
\textbf{Uniform (Early Bias)} & \textbf{[3, 7, 11, 15]} & \textbf{[2, 5, 8, 11]} & \textbf{34.59} & \textbf{0.2746} \\
\bottomrule
\vspace{-25pt}
\end{tabular}
\end{table}
To understand the impact of the interaction module's placement, we conducted an ablation study by inserting four BridgeDiT Blocks at different stages within the dual-tower architecture. We evaluated five distinct placement strategies: concentrating the blocks in the early, middle, or late layers, as well as two uniform distribution strategies.

The results, presented in Table~\ref{tab:placement_ablation}, reveal a clear trend. The Uniform (Early Bias) strategy, where blocks are inserted uniformly across the first half of the network layers, yields the best performance on both the ImageBind (VA-IB) and AV-Align metrics. Performance is strongest when interaction occurs in the early-to-mid layers, as seen in the ``Middle Layers" and ``Uniform" configurations. Conversely, concentrating the interaction exclusively in the deepest, final layers (``Late Layers") results in a significant degradation of performance. This suggests that for achieving robust audio-visual synchronization, the most critical feature exchange occurs at the early and intermediate representational stages. We hypothesize that these layers contain the optimal balance of detailed spatial-temporal information (from early layers) and abstract semantic concepts (from middle layers). Relying only on the highly abstract features from the final layers is insufficient for the precise alignment required for the T2SV task.



\section{Additional Experimental Results}
\label{sec:appendix_additional}

To further validate our method, we report three additional studies:
(i) cross-backbone generalization together with comparisons against recent
baselines and an out-of-distribution (OOD) evaluation
(Sec.~\ref{subsec:crossbackbone}); and
(ii) the effect of LLM scale in the CRR caption framework, including its
inference-time cost (Sec.~\ref{subsec:llmscale}).

\subsection{Cross-Backbone Generalization, Additional Baselines, and OOD Evaluation}
\label{subsec:crossbackbone}

Table~\ref{tab:appendix_crossbackbone} jointly addresses three questions on
the AVSync15 test set.

\noindent\textbf{(i) Cross-backbone generalization.}
We replace the default Wan~/~SDA pair with a different backbone pair,
SANA-Video~\cite{chen2025sanavideo} and AudioLDM~\cite{liu2024audioldm}. Adding our DCA
interaction (\emph{SANA+AudioLDM (DCA)}) consistently improves every metric
over the non-interacting separate baseline
(\emph{SANA+AudioLDM (separate)}), e.g., FAD $17.09 \!\rightarrow\! 9.46$ and
AV-Align $0.193 \!\rightarrow\! 0.256$. This confirms that DCA is not tied to a
specific backbone pair and generalizes across architectures.

\noindent\textbf{(ii) Additional baselines.}
We compare against recent cascaded V2A pipelines (Wan~+~ThinkSound~\cite{liu2025thinksoundchainofthoughtreasoningmultimodal},
Wan~+~PrismAudio~\cite{liu2025prismaudiodecomposedchainofthoughtsmultidimensional}) and recent unified audio-video models
(Ovi~\cite{low2025ovi}, LTX-2~\cite{hacohen2026ltx}). BridgeDiT outperforms the cascaded
pipelines on most quality and alignment metrics. LTX-2 attains a lower FVD and
CLIPSIM, which we attribute to its substantially larger and stronger video
backbone; moreover, both LTX-2 and Ovi primarily target speech and dialogue
rather than the foley setting we focus on, and are therefore reported here for
completeness rather than as direct, fairness-controlled comparisons.

\noindent\textbf{(iii) OOD evaluation.}
\emph{BridgeDiT (VGG-SS, OOD)} is trained on VGGSound-SS and evaluated on
AVSync15 without any in-domain training. Despite the domain shift, it still
outperforms Wan~+~ThinkSound and remains competitive with Wan~+~PrismAudio,
indicating that our gains stem from the method itself rather than from
per-dataset tuning.

\begin{table}[t]
\centering
\caption{Cross-backbone generalization, additional baselines, and OOD
evaluation on the AVSync15 dataset. \textbf{Best} and \underline{second-best}
are highlighted. LTX-2 and Ovi use larger and/or speech-oriented backbones and
are listed for completeness. ``OOD'' denotes BridgeDiT trained on VGGSound-SS
and tested on AVSync15.}
\vspace*{-8pt}
\renewcommand{\arraystretch}{1.2}
\resizebox{\textwidth}{!}{%
\begin{tabular}{ll|cc|cc|cc}
\toprule[1.5pt]
\textbf{Setting} & \textbf{Method} & FVD$\downarrow$ & FAD$\downarrow$ & CLIPSIM$\uparrow$ & CLAP$\uparrow$ & VA-IB$\uparrow$ & AV-Align$\uparrow$ \\
\hline
\multirow{2}{*}{Cross-Backbone (Ours)}
& SANA + AudioLDM (separate) & 1033.39 & 17.09 & 27.95 & 29.66 & 23.39 & 0.193 \\
& SANA + AudioLDM (DCA)      & 906.29  & 9.46  & 28.46 & 32.67 & 31.85 & 0.256 \\
\hline
\multirow{2}{*}{Cascaded V2A}
& Wan + ThinkSound  & 828.33 & 17.31 & 28.12 & 30.92 & 20.70 & 0.169 \\
& Wan + PrismAudio  & 828.33 & 9.79  & 28.12 & 33.53 & 25.25 & 0.233 \\
\hline
\multirow{2}{*}{Unified (larger/speech)}
& Ovi   & 790.94 & 5.58  & 26.38 & 34.12 & \underline{31.93} & 0.253 \\
& LTX-2 & \textbf{701.21} & 10.92 & \textbf{30.37} & 32.13 & 28.78 & \underline{0.258} \\
\hline
OOD & BridgeDiT (VGG-SS, OOD) & 826.50 & 12.95 & 28.34 & \underline{34.47} & 26.27 & 0.239 \\
\hline
\rowcolor{mypalelue}
\cellcolor{white} Default & \textbf{BridgeDiT (Ours)} & \underline{765.74} & \textbf{5.34} & \underline{28.52} & \textbf{35.95} & \textbf{34.59} & \textbf{0.275} \\
\bottomrule
\end{tabular}%
}
\label{tab:appendix_crossbackbone}
\end{table}

\subsection{Effect of LLM Scale in the CRR Framework}
\label{subsec:llmscale}

Our CRR caption framework uses Qwen2.5-72B for both the Semantic Checker and
the Cross-Modal Rewriter. To assess whether the gains come from the framework
design or merely from model scale, Table~\ref{tab:appendix_llmscale} replaces
the 72B model with a 7B counterpart and additionally reports the per-video
prompt-expansion latency. The 7B variant remains effective across all metrics,
indicating that the improvements primarily arise from the Semantic Anchor
bottleneck and cross-modal rewriting rather than from LLM scale alone. The 72B
model yields the best quality at the cost of higher latency. Importantly, for
training this prompt expansion is performed \emph{offline} and the resulting
captions can be cached and reused, so the practical overhead is minimal.

\begin{table}[t]
\centering
\caption{Effect of replacing the 72B Checker/Rewriter with a 7B model on
AVSync15, including per-video prompt-expansion latency. \textbf{Best} per
column is highlighted.}
\vspace*{-8pt}
\renewcommand{\arraystretch}{1.2}
\resizebox{\textwidth}{!}{%
\begin{tabular}{l|cc|cc|cc|c}
\toprule[1.5pt]
\textbf{Method} & FVD$\downarrow$ & FAD$\downarrow$ & CLIPSIM$\uparrow$ & CLAP$\uparrow$ & VA-IB$\uparrow$ & AV-Align$\uparrow$ & Latency/video (s)$\downarrow$ \\
\hline
CRR (7B)              & 787.93 & 8.26 & 27.37 & 31.78 & 32.34 & 0.258 & \textbf{8.07} \\
\rowcolor{mypalelue}
\textbf{CRR (Ours, 72B)} & \textbf{765.74} & \textbf{5.34} & \textbf{28.52} & \textbf{35.95} & \textbf{34.59} & \textbf{0.275} & 30.15 \\
\bottomrule
\end{tabular}%
}
\label{tab:appendix_llmscale}
\end{table}

%

\section{Future Work}
\label{app:future}
These limitations pave the way for several exciting future directions. A crucial step is the collection of larger, higher-quality audio-visual datasets, coupled with more efficient data processing pipelines for cleaning, filtering, and captioning. Building on our architecture, we plan to extend BridgeDiT to support speech and music. This will involve incorporating specialized modules for lip-synchronization and developing techniques to capture the rhythm and mood of musical inputs. Moreover, we are interested in exploring post-training refinement techniques. For instance, applying Reinforcement Learning with Human Feedback (RLHF), with rewards specifically designed to enhance audio-visual synchronization, could further improve the model's temporal and semantic coherence. We believe these future steps will continue to advance the field towards the generation of truly holistic and synchronized multi-sensory experiences.

\end{document}